\documentclass[10pt,twocolumn,letterpaper]{article}

\usepackage[pagenumbers]{iccv} 

\definecolor{iccvblue}{rgb}{0.21,0.49,0.74}
\usepackage[pagebackref,breaklinks,colorlinks,allcolors=iccvblue]{hyperref}

\usepackage{graphicx}
\usepackage{amsmath}
\usepackage{amssymb}
\usepackage{booktabs}
\usepackage[dvipsnames]{xcolor}
\usepackage{multirow}
\usepackage{pifont}
\usepackage{comment}
\usepackage{tikz} 
\usepackage{pgfplots} 
\pgfplotsset{compat=1.18}

\usepackage[capitalize]{cleveref}
\crefname{section}{Sec.}{Secs.}
\Crefname{section}{Section}{Sections}
\Crefname{table}{Table}{Tables}
\crefname{table}{Tab.}{Tabs.}

\usepackage{sidecap} 

\def\paperID{8345} 
\def\confName{ICCV}
\def\confYear{2025}

\def\showcomments{0}
\newcommand\ifcom[3]{
\ifx\showcomments\undefined

\else
  \if\showcomments 1
  \textcolor{#1}{\small\bf\sf [#2: #3]}
  \else
  {}
  \fi
\fi
}

\definecolor{darkgreen}{rgb}{0.0,0.6,0.0}

\definecolor{cyan}{rgb}{0.45,0.87,0.95}
\definecolor{orange}{rgb}{0.95,0.8,0.6}
\definecolor{britishracinggreen}{rgb}{0.0, 0.26, 0.15}
\definecolor{cadmiumgreen}{rgb}{0.0, 0.42, 0.24}

\newcommand\eqvspace{\vspace{-0.0cm}}

\definecolor{citecolor}{RGB}{34,139,34}
\definecolor{lightred}{RGB}{241,140,142}
\title{\vspace{-8pt}Teaching VLMs to Localize Specific Objects from In-context Examples}
\author{
    Sivan Doveh*$^{2}$\quad
    Nimrod Shabtay*$^{1,3}$\quad
    Wei Lin$^{4}$\\
    Eli Schwartz$^{1}$\quad
    Hilde Kuehne$^{1,5}$\quad
    Raja Giryes$^{3}$\quad
    Rogerio Feris$^{6}$\\
    Leonid Karlinsky$^{6}$\quad
    James Glass$^{7}$\quad
    Assaf Arbelle$^{1}$\quad
    Shimon Ullman$^{2}$\quad
    M. Jehanzeb Mirza$^{7}$
    \\
    \small{$^{1}$IBM Research\quad
    $^{2}$Weizmann Institute of Science\quad
    $^{3}$Tel Aviv University,}\\
    \small{$^{4}$JKU Linz\quad
    $^{5}$Tuebingen AI Center\quad
    $^{6}$MIT-IBM\quad
    $^{7}$MIT CSAIL}}
\begin{document}
\twocolumn[{
\renewcommand\twocolumn[1][]{#1}
\maketitle
\centering
\vspace{-20pt}
\includegraphics[width=0.75\textwidth]{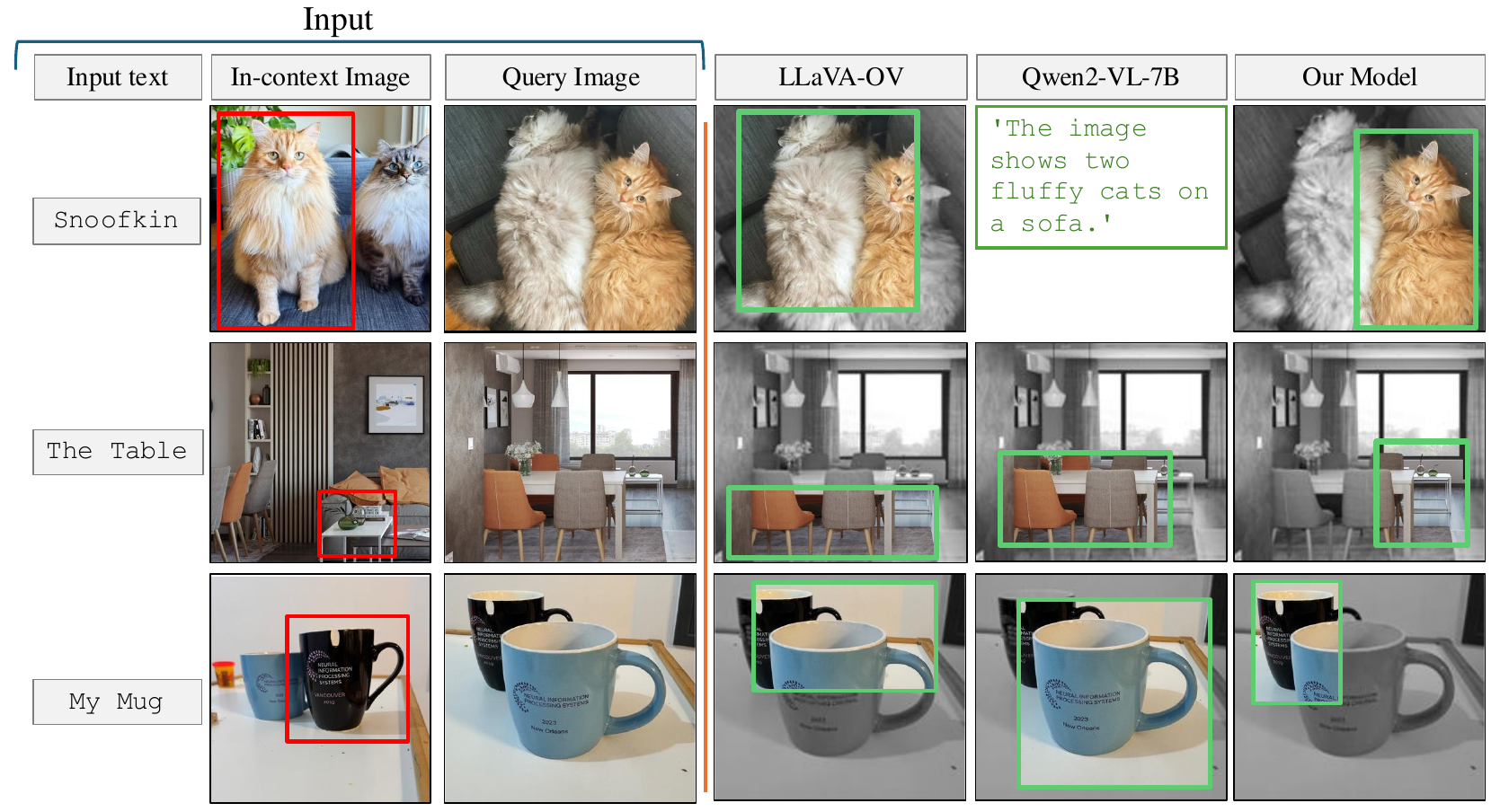}
\captionof{figure}{\textbf{In-context personalized localization} involves localizing object instances present in a scene (or query image) similar to the object presented as an in-context example.
In this setting, the input to the model is a \texttt{category name}, \texttt{in-context image}, \texttt{bounding box coordinates} (not shown in this figure), and a \texttt{query image}. The model is tasked with localizing the \textit{same} category of interest (presented as an in-context example) in the \texttt{query image}. Here, we visualize a few inputs and outputs from various VLMs highlighting that our fine-tuned model better captures the information in the in-context image.}
\label{fig:teaser}
\vspace{.2cm}
}]
\newcommand{\minisection}[1]{\noindent{\textbf{#1}.}}
\maketitle
        
\newcommand{\method}{IPLoc\xspace}
 
\newcommand{\llava}{LLaVA\xspace}
\newcommand{\llama}{Llama\xspace}

\newcommand{\ClipEnc}{\theta}
\newcommand{\vis}{\phi}  
\newcommand{\txt}{\psi}  
\newcommand{\visEnc}{\vis} 
\newcommand{\txtEnc}{\txt} 

\newcommand{\txtEmb}{t} 
\newcommand{\txtEmbNew}{\txtEmb_{\hat{c}}} 

\newcommand{\imgEmb}{\vis} 

\newcommand{\cls}{c} 
\newcommand{\ClassNames}{C} 

\newcommand{\img}{x} 

\newcommand{\prompt}{p} 

\newcommand{\simil}{\cos} 

\newcommand{\likel}{l} 

\newcommand{\LSCE}{\mathcal{L}_\textsc{SCE}}

\newif\ifdraft
\draftfalse
\drafttrue 
\ifdraft
 \newcommand{\MK}[1]{{\color{magenta}{\bf MK: #1}}}
  \newcommand{\JM}[1]{{\color{blue}{\bf JM: #1}}}
\newcommand{\LK}[1]{{\color{red}{\bf LK: #1}}}

 \newcommand{\mk}[1]{{\color{magenta} #1}}
\else
 \newcommand{\MK}[1]{}
 \newcommand{\mk}[1]{#1}
\fi

\newcommand{\quotebox}[1]{\begin{center}\fcolorbox{white}{blue!15!gray!15}{\begin{minipage}{1\linewidth}\vspace{1pt}\center\begin{minipage}{1\linewidth}{\space\Huge``}{#1}{\hspace{1.5em}\break\null\Huge\hfill''}\end{minipage}\smallbreak\end{minipage}}\end{center}}


 \newcommand{\tick}{\ding{51}}
  \newcommand{\cross}{\ding{55}}

  \definecolor{mycolor}{RGB}{176,64,35}

    \definecolor{llmcolor}{RGB}{255,18,18}
\newcommand{\flamingo}[1]{\textcolor{mycolor}{#1}}

\newcommand{\llmresponse}[1]{\textcolor{llmcolor}{#1}}

\newcommand\figcapvspace{\vspace{-2cm}}
\newcommand{\gen}[1]{\mathcal{G}(#1)}

\newcommand{\emb}[1]{\texttt{emb}(#1)}

\newcommand\secv{\vspace{0cm}}
\newcommand\subsecvt{\vspace{0cm}}
\newcommand\subsecvb{\vspace{0cm}}
\newcommand\secvb{\vspace{0cm}}
\newcommand\figvb{\vspace{0cm}}
\newcommand\figvt{\vspace{0cm}}
\newcommand\tabvb{\vspace{0cm}}
\newcommand\tabvcap{\vspace{0cm}}
\newcommand\figvcap{\vspace{0cm}}

\newcommand\eqvt{\vspace{-0.0cm}}
\newcommand\eqvb{\vspace{0cm}}
\newcommand{\footnoteref}[1]{\textsuperscript{\ref{#1}}}

\def\eg{\emph{e.g.,}\xspace} 
\def\Eg{\emph{E.g}\onedot}
\def\ie{\emph{i.e.,}\xspace} 
\def\Ie{\emph{I.e}\onedot}
\def\cf{\emph{c.f.,}\xspace} 
\def\Cf{\emph{Cf}\onedot}
\def\etc{\emph{etc}\onedot} 
\def\vs{\emph{vs}\onedot}
\def\wrt{w.r.t\onedot} 
\def\dof{d.o.f\onedot}
\def\iid{i.i.d\onedot} 
\def\wolog{w.l.o.g\onedot}
\def\etal{\emph{et al.}}

\begin{abstract}
\vspace{-0.7cm}

Vision-Language Models (VLMs) have shown remarkable capabilities across diverse visual tasks, including image recognition, video understanding, and Visual Question Answering (VQA) when explicitly trained for these tasks. 
Despite these advances, we find that present-day VLMs (including the proprietary GPT-4o) lack a fundamental cognitive ability: learning to localize specific objects in a scene by taking into account the \emph{context}.
In this work, we focus on the task of few-shot personalized localization, where a model is given a small set of annotated images (in-context examples) -- each with a category label and bounding box -- and is tasked with localizing the same object type in a query image. 
Personalized localization can be particularly important in cases of ambiguity of several related objects that can respond to a text or an object that is hard to describe with words.\footnote{Code: \url{https://github.com/SivanDoveh/IPLoc}} 
To provoke personalized localization abilities in models, we present a data-centric solution that fine-tunes them using carefully curated data from video object tracking datasets. 
By leveraging sequences of frames tracking the same object across multiple shots, we simulate instruction-tuning dialogues that promote context awareness. 
To reinforce this, we introduce a novel regularization technique that replaces object labels with pseudo-names, ensuring the model relies on visual context rather than prior knowledge. 
Our method significantly enhances the few-shot localization performance of recent VLMs ranging from 7B to 72B in size, without sacrificing generalization, as demonstrated on several benchmarks tailored towards evaluating personalized localization abilities. 
This work is the first to explore and benchmark personalized few-shot localization for VLMs -- exposing critical weaknesses in present-day VLMs, and laying a foundation for future research in context-driven vision-language applications.
\end{abstract}
\section{Introduction}
\label{sec:intro}

Present-day Vision-Language Models (VLMs)~\cite{clip, clippr, llava, llava+, idefics3, qwen-vl}, which aim to emulate human learning processes, have achieved unprecedented performance across tasks such as image recognition~\cite{clip, mpvr}, video understanding~\cite{lin2023video}, and Visual Question Answering (VQA)~\cite{llava-next, qwen-vl}. 
Despite strong capabilities in certain domains, VLMs still lack a fundamental aspect of human cognition: the ability to learn from \emph{context}.
This type of learning, known as in-context learning (ICL) in large language models (LLMs) and VLMs, refers to the model’s ability to draw on cues from in-context examples to infer relevant information in a query. 
It is surprising that VLMs, despite combining a robust vision encoder (\eg~the CLIP~\cite{clip} vision encoder) with a powerful language decoder (\eg~\llama~\cite{dubey2024llama}), struggle with in-context inference on visual tasks. 
While LLMs excel at ICL~\cite{icl, cot, treeofthought}, this ability does not readily transfer to multimodal contexts, rendering VLMs unable to mimic this essential aspect of human cognition.
One possible reason for this gap is the nature of the data used for instruction-tuning VLMs. 
These data, often curated from LLMs~\cite{gpt3} and based on general-purpose image collections (\eg~MS-COCO~\cite{mscoco}), lack semantically correlated instructions that focus on contextual learning. 
Although this approach enhances the model’s reasoning about specific instances, it does not encourage the model to focus on context for generalized learning.

Recent works have begun addressing ICL limitations in VLMs for tasks like few-shot object classification and VQA~\cite{llava_icl,task-vec,tv_mm}. 
Despite their success in these tasks, these approaches have not been extended to other tasks like object localization, and, to our knowledge, no work has focused specifically on few-shot localization.
To fill this gap, in this work we focus on the task of few-shot personalized object localization. 
In this task, given a set of $N$ example images annotated with a category label (\eg~\texttt{snoofkin} -- the name of your cat) and bounding-box coordinates, the model’s objective is to localize the same type of object (\ie~\texttt{snoofkin}) in a query image (acquired from a different scene).
This task is particularly valuable in scenarios where reference images are scarce, yet efficient localization is needed across large datasets or video streams. 
It also helps disambiguate objects with close semantic resemblance.
We highlight the setting and some of these ambiguous cases (in addition to model outputs) in Figure~\ref{fig:teaser}.

We enhance in-context personalized localization abilities in VLMs using a data-centric approach, termed as \method. 
\method renders VLMs more context-aware by fine-tuning off-the-shelf VLMs on data carefully harnessed from video object tracking datasets. 
These datasets provide an ideal training environment, as they track the same object instance across multiple frames (stimulating the \emph{personalization} aspect). 
Specifically, we employ three large-scale video tracking datasets~\cite{tao, lasot, got} and sample frames to construct structured conversations between \textit{user} and \textit{assistant}, in line with instruction-tuning formats. 
Furthermore, these conversations are organized by the \emph{number of shots}, 
to train these VLMs to utilize contextual information from in-context examples effectively.
Further, to prevent models from relying on pre-trained object knowledge, our \method introduces a simple yet effective regularization technique: renaming object categories with \emph{pseudo-names} (\eg renaming “Airplane” as “Elizabeth”) to ensure that models focus on contextual cues for object identification, and also induces the personalization aspect as it is a specific named-entity and not a generic object category. 
After creating these personalized few-shot dialogues, we instruction-tune the VLMs in a parameter-efficient manner to preserve generalization.
We extensively evaluate state-of-the-art proprietary (\eg~GPT-4o~\cite{openai2023gpt4}) and open VLMs (\eg~Qwen-VL~\cite{qwen2vl}) of various sizes, highlighting critical weaknesses of these models for in-context localization. 


Our contributions can be summarized as follows: 
\begin{itemize} \item We are the first to explore, propose benchmarks, and extensively study personalized ICL abilities in modern VLMs (including the proprietary GPT-4o) and find that these models struggle critically in this task. 
\item We propose a data-centric approach that can be bootstraped with any open VLM and utilizes multiple object-tracking datasets to generate instruction-tuning conversations and fine-tunes them for improved context awareness for the task of personalized few-shot object localization. 
\item We show the generalization of our method on various model families and sizes (up to 72B parameters), and find that our method prevents the VLMs from losing previously learned abilities during the pre-training.  \end{itemize}

\section{Related Work}
\label{sec:related_work}
Our work is closely related to large-scale vision-language foundation models and approaches that enhance the in-context learning abilities of these models. 

\paragraph{Vision-Language Foundation Models} can typically be divided into two distinct categories. 
One category of methods usually relies on two encoders (vision and text encoders), which are trained in a contrastive manner on large-scale paired image-text data that is scraped from the web. 
These models show strong retrieval abilities and flourish at discriminative tasks like image recognition. 
Some popular methods which follow this line of work involve CLIP~\cite{clip}, ALIGN~\cite{align}, OpenCLIP~\cite{openclip}, SigLIP~\cite{siglip}, and MetaCLIP~\cite{metaclip}. 
Some methods,~\eg~\cite{mpvr,lafter,svlc,dac,maxi,tap,glov} further build upon these contrastively pre-trained models to improve them for specific downstream tasks like video action recognition, compositionality reasoning,~\etc.
The other category of methods builds upon two components: a pre-trained vision backbone (usually a contrastively pre-trained CLIP~\cite{clip} vision encoder) and a decoder (usually an LLM~\cite{vicuna2023,dubey2024llama}).
These two components are grounded with a projection from the vision embedding space to the LLM input embedding space,~\eg through a projector. 
This projection equips the LLM to process the visual tokens and perform open-ended visual reasoning tasks. 
Some representative approaches belonging to this category include BLIP~\cite{blip2}, Instruct-BLIP~\cite{instructblip}, MiniGPT~\cite{minigpt, minigptv2}, Qwen-VL~\cite{qwen-vl, qwen2vl}, Idefics~\cite{idefics3}, and the \llava family of models~\cite{llava,llava-ov}.
These models have shown unprecedented performance gains on many reasoning tasks, like VQA, scene understanding, compositionality reasoning,~\etc. 
However, it is reported that these models still somewhat lack in their abilities to learn from context~\cite{llava_icl, task-vec}. 
In this work, our goal is to render these VLMs more context-aware. 
To this end, we focus on the task of personalized localization of objects, where the model is required to localize an object in a query image while taking cues from the images (with annotations) provided as context. 

\paragraph{In-context Learning for VLMs:} 
The ability of models to solve novel tasks by consuming a few demonstrations of the downstream task of interest has been formalized as in-context learning (ICL)~\cite{icl}.
In the natural language processing (NLP) domain, several methods have demonstrated strong in-context learning abilities for various downstream tasks of interest.  
Notably,~Brown~\etal~\cite{brown2020language} popularized few-shot learning for LLMs.
Similarly, other methods~\cite{cot, treeofthought} also present different ways to elicit the in-context learning abilities in LLMs. 
For VLMs, prominent works like Flamingo~\cite{flamingo} showed that few-shot ICL can scale up to large-scale vision language models by training on a large corpus of interleaved image-text data. 
Yet, in recent VLMs, \eg~\llava~\cite{llava}, which inherit a strong decoder (an LLM), usually the ICL capabilities are not directly transferred to the VLMs. 
This could be due to the generic instruction tuning data which does not force the model to focus on the context, but instead teaches the model strong reasoning abilities about the current data instance only.

Some methods like Emu2~\cite{emu2}, Idefics~\cite{idefics,idefics3}, MM-ICL~\cite{mmicl}, and Qwen-VL~\cite{qwen-vl} focus on training models to specifically improve the multi-modal ICL abilities. 
They commonly do so by scaling the model size and designing specific training data. 
On the other hand, some approaches improve the ICL abilities of VLMs in a post hoc manner. 
Specifically, \llava-ICL~\cite{llava_icl} finetunes off-the-shelf VLMs in a parameter-efficient manner to improve their few-shot classification and VQA abilities, on thematically consistent data.
MTV~\cite{task-vec} focuses on mitigating the limited context length issues in VLMs by introducing a training-free approach to find task vectors to feed multiple in-context shots to the network, which can even exceed the true context length of the network. 
Many of these VLMs primarily focus on improving few-shot classification or VQA tasks. 

Through extensive experimentation, we find that many of these VLMs perform underwhelmingly when evaluated on the task of few-shot localization and struggle even more for~\emph{personalized} object localization, where the goal is to localize the same instance of the object in the query data. 
Our \method takes steps towards improve these models for the task of personalized object localization by leveraging data from video object tracking datasets which help us to harness instruction-tuning conversations for fine-tuning these models. 
This data helps the models to become~\emph{context-aware} and induce the aspect of personalization, since the conversations are focused on a single object in multiple different frames (sampled at different time steps).
Our parameter-efficient fine-tuning strategy shows impressive gains over the baseline models without the loss of generalization.  


\begin{figure*}[t!]
    \centering
    \includegraphics[width=0.8\textwidth]{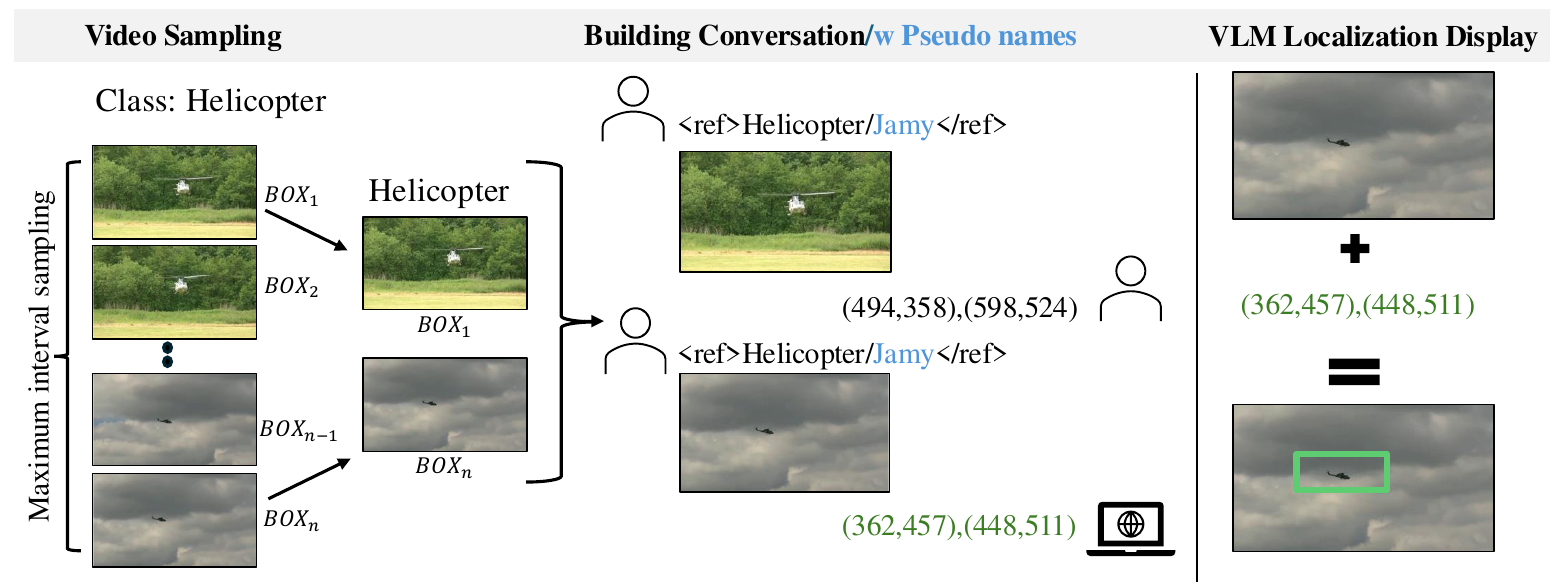} 
    \tabvcap
    \caption{\textbf{Overview of data creation and conversation format}. To instill few-shot personalized localization abilities in VLMs, our \method creates multi-modal conversations by harnessing data from multiple video object tracking datasets. For semantic coherence, focus on personalization and stronger contextual awareness, we create these conversations by sampling frames from the same video, tracking a particular object of interest, and enhancing the training data by extending the conversations by replacing the \emph{true} category name with~\emph{pseudo names}. These conversations are later employed to induce contextual awareness in VLMs.}
    \label{fig:method}
\figvcap

\end{figure*}

\section{\method: In-context Personalized Localization}
\label{sec:method}

In this work, we focus on leveraging existing pre-trained (encoder-decoder) VLMs to make them context aware, particularly for the task of personalized few-shot object localization. 
To this end, we harness instruction-tuning data from video object tracking datasets because such data usually tracks the same object over multiple frames, thus, instilling a notion of personalization in the instruction-tuning conversations.
In the following, we explain the details of the proposed approach which enables models to focus on contextual cues. 
For ease of assimilation, we divide the explanation into $3$ distinct parts.
In Section~\ref{subsec:icl-inst-tuning} we explain how we construct the instruction tuning conversations, then in Section~\ref{subsec:data-mixes} we outline our data curation strategy and finally in Section~\ref{subsec:fine-tuning} we conclude with an explanation of the choice of the fine-tuning strategy we employ.

\subsection{ICL Instruction Tuning Conversations}
\label{subsec:icl-inst-tuning}
We aim to build instruction-tuning conversations that can be used to fine-tune existing VLMs such that they can become context-aware for the task of localization. 
Our ICL instruction-tuning format intends to provide an improvement to the general instruction-tuning format of~\cite{qwen2vl, llava-ov} that is not able to process multi-modal ICL instructions.
These VLMs usually 
align modality-specific encoders (\eg~visual, audio, speech) with a capable decoder (\ie~an LLM) via multi-modal instruction tuning combined with a structured training curriculum. 
Our work builds upon existing pre-trained multi-modal alignment architectures where a pre-trained modality encoder $\mathcal{E}$, a modality projector $\mathcal{P}$, and an LLM decoder $\mathcal{D}$ are integrated, extending prior methods like \cite{qwen-vl}.

Our proposed instruction-tuning data format typically comprises multi-modal conversations, where a simulated \textit{user} interacts with an \textit{assistant}.
In our case, each conversation consists of multiple messages, where each message in the conversation follows the following format:
\small
\[
\text{Message: } \texttt{<image><ref>category</ref>}
\]
\normalsize
accompanied by bounding box coordinates:
\small
\[
((x_{min}, y_{min}), (x_{max}, y_{max})).
\]
\normalsize
Specifically, each message consists of an image, the corresponding category name (\eg~\texttt{caterpillar}), and the bounding box localizing the object of interest. 



To obtain conversations for few-shot localization, we structure them for in-context learning with coherent localization instructions. 
Each few-shot (\eg $\text{n}$-shot) localization setup follows this format:

\small
\begin{align*}
\eqvspace
&\text{User: } S_1^1 \, \texttt{<image><ref>Category</ref>} \, S_1^2 \\
&\quad (x_{1,1}, y_{1,1}), (x_{1,2}, y_{1,2}) \\
&\text{User: } S_2^1 \, \texttt{<image><ref>Category</ref>} \, S_2^2 \\
&\quad  (x_{2,1}, y_{2,1}), (x_{2,2}, y_{2,2}) \\
&\vdots \\
&\text{User: } S_n^1 \, \texttt{<image><ref>Category</ref>} \, S_n^2 \\
&\quad  (x_{n,1}, y_{n,1}), (x_{n,2}, y_{n,2}) \\
&\text{User: } \texttt{<image><ref>Prediction</ref>} \\
& \text{Assistant: } (x_{\text{n+1},1}, y_{\text{n+1},1}), (x_{\text{n+1},2}, y_{\text{n+1},2})
\eqvspace
\end{align*}
\normalsize

In this setup, each turn labeled as “User” includes an input comprising contextual text $S_j^1$ preceding an image reference with a known category label, denoted as \texttt{<image><ref>Category</ref>}, followed by additional contextual text $S_j^2$. Each user input also specifies a bounding box with coordinates \((x_{j,1}, y_{j,1}), (x_{j,2}, y_{j,2})\), which delineate the precise location of the labeled object within the image.
The assistant then responds with its prediction, which is structured as \texttt{<image><ref>Prediction</ref>}, where it provides a category label based on the context provided by the preceding user examples and the bounding box coordinates for the category of interest. 
One such conversation is also outlined in Figure~\ref{fig:method}.

This structured format allows the model to learn to associate provided bounding boxes with corresponding categories across multiple in-context examples and to output the category associated with the specific location (delineated by the bounding box coordinates provided) in the query image. 
Using input masking for human turns, left-attentive causal language modeling (CLM), and the designed ICL format, the model functions as an ``any-shot" trainer: the first shot acts as a zero-shot instruction replay, and each subsequent shot progressively adapts the model for few-shot localization tasks by training on $1, \cdots, (i-1)$ previous examples in its attended context.


\subsection{Data Mixes} \label{subsec:data-mixes}

Curating fine-tuning data carefully is central to our methodology for making current VLMs more context-aware in the task of few-shot personalized localization. 
A straightforward approach to constructing conversational fine-tuning data is to leverage established object detection and localization datasets, such as MS-COCO~\cite{mscoco}, and build dialogues as outlined in Section~\ref{subsec:icl-inst-tuning}. 
This approach, however, often results in generic instruction-tuning data akin to the popularized methods of~\llava~\cite{llava}. 
As recent research has indicated~\cite{llava_icl}, these generic data mixes can hinder VLMs from effectively attending to contextual cues essential for nuanced and situational understanding.

To address this, our data design prioritizes instruction-tuning conversations that are semantically coherent and focused on personalization (\eg~relating to the same object in the conversations) and structured to enable the few-shot learning capabilities required by our models. 
Focusing on the specific needs of our task, we aim to design instruction tuning data that promotes~\emph{personalized} retrieval, where the model learns to focus on unique objects based on the provided context as few-shot examples.
With the overarching goal of training VLMs to hone in on contextual cues for few-shot object localization, we select three prominent single- and multi-object tracking datasets as our foundation for conversation creation between a \emph{user} and an \emph{assistant} (\cf Section~\ref{subsec:icl-inst-tuning}):

\begin{itemize} \item \textbf{Tracking Any Object (TAO)}~\cite{tao} is a large-scale multi-object tracking dataset comprising $839$ categories and $2907$ high-resolution videos. 
TAO’s diversity and category breadth provide a strong foundation for building complex, multi-faceted instruction-tuning conversations. \item \textbf{Large-scale Single Object Tracking (LaSOT)}~\cite{lasot} includes $1400$ sequences and over $3.5$ million frames, with videos averaging $2500$ frames each. 
This dataset’s extended, high-frame sequences make it well-suited for developing long-term instruction-tuning dialogues focused on a single, persistent object. 
\item \textbf{Generic Object Tracking Benchmark (GOT)}~\cite{got} offers an extensive set of $10000$ video segments of real-world moving objects, providing a rich dataset for exploring variations in object movement and context. \end{itemize}

These datasets, with their extensive tracking sequences of diverse real-world objects, provide an ideal testing ground for developing personalized few-shot tracking conversations. Our fine-tuning conversations emphasize data centered on a single object being tracked across frames, with a deliberate design choice for maximum interval sampling. 
This approach introduces variation in the object’s position and appearance (\eg~rotation, lighting, background~\etc) across frames, thereby increasing the challenge for the model and encouraging it to consistently localize the specific object of interest despite diverse contextual shifts.
An overview of the data curation is provided in Figure~\ref{fig:method}.

Through this design, we generate fine-tuning data composed of multiple shots (\eg~1-8) in which each conversation focuses on localizing a single object of interest within a single video sequence. 
This structure reinforces both personalization (due to the recurring task of localizing the same object) and semantic coherence (by sampling frames from the same video sequence). 
To validate our approach, we conduct an ablation study comparing our proposed data curation strategy with an alternative where conversations are generated by sampling data for a particular category from unrelated video sequences. 
As shown in Table~\ref{tab:semantically_coherent_data}, our data curation approach significantly improves model performance, underscoring the effectiveness of structured sampling for enhancing context-aware localization.

To strongly bias the model toward focusing on contextual cues, rather than relying solely on knowledge acquired during large-scale pre-training, we propose a simple regularization technique to enhance the training data. 
Specifically, we replace true class names with \emph{pseudo-names} (\eg replacing “Jaguar” with “John”). 
This regularization encourages the model to depend less on its pre-trained knowledge of the category “Jaguar” when outputting localization coordinates during the \emph{assistant} turn, and instead biases it to rely on the few-shot in-context examples, which refer to the pseudo-name “John”.
The main results listed in Table~\ref{tab:main-results} demonstrate the effectiveness of this regularization, indicating substantial improvements in model performance. 

In summary, our fine-tuning dataset consists of a combination of data mixes from three video object-tracking datasets~\cite{tao, lasot, got}, with conversations spanning between $1$-$8$ shots, as well as data generated by replacing the \emph{true} category names with \emph{pseudo} category names. 
In the next section, we outline how this large corpus\footnote{The entire dataset with data mixes will be released upon acceptance.} is employed to enhance VLM performance for personalized ICL.

\subsection{Fine-tuning} \label{subsec:fine-tuning}

To fully leverage our curated shot-based conversational data described in Section~\ref{subsec:data-mixes}, we employ a targeted fine-tuning strategy designed to enhance VLM performance for few-shot personalized localization. 
To recap, each fine-tuning conversation comprises multiple shots (1–8) sampled from within a single video sequence coupled with the \emph{pseudo} category name regularized data, with a consistent focus on localizing a single object across frames. 
While there exist multiple approaches to fine-tune VLMs, including full-model adaptation and adapter-based tuning, we adopt LoRA~\cite{lora} as our fine-tuning method of choice. 
By embedding these conversations within LoRA’s compact and memory-efficient fine-tuning schema, we encourage the model to develop a more refined contextual awareness specific to the personalized few-shot localization task, without compromising the broad knowledge encoded in the VLM's pre-trained parameters.
The learning objective is the standard language modeling (next-token-prediction) loss and it guides the model in accurately generating localization coordinates for the target object within the frame of a shot during the \emph{assistant} turn of the conversation (\cf~Section~\ref{subsec:icl-inst-tuning}), biasing the model to focus on contextual cues.
During each conversational exchange, the \emph{assistant} learns to predict precise bounding-box coordinates by conditioning its responses on the contextual cues provided by the preceding \emph{user} examples. 
This setup encourages the model to rely on the in-context examples, adapting its language-based understanding to capture spatial relationships within each conversation (meticulously curated from a single video).

LoRA limits the extent of parameter adjustments thereby reducing the risk of overfitting to the specific localization task.
This is further highlighted in an ablation listed in Table~\ref{tab:retention_generalization}, where we compare fully fine-tuned models and LoRA fine-tuned models on the standard VLM benchmarks~\cite{gqa,fang2021seed,pope} and find that our fine-tuning strategy maintains the generalization abilities of the VLM by only incurring a minimal performance penalty of $\sim1\%$ as compared to the base pre-trained model.

\section{Experimental Results}
\label{sec:experiments}
In this section, we first list all the evaluation settings, including datasets, implementation details, and metrics. Then, we provide a detailed discussion of the main results and conclude by extensively ablating our~\method.  

\subsection{Evaluations Settings}
\label{subsec:eval_settings}
\paragraph{Datasets:} To generate the multi-modal conversations (\cf Section~\ref{subsec:icl-inst-tuning}) for making the VLMs context-aware we employ $3$ large scale datasets: TAO~\cite{tao}, LASOT~\cite{lasot} and GOT~\cite{got}. 
These datasets provide bounding box annotations and the category label and we leverage them to construct the conversations for fine-tuning, as explained in Section~\ref{subsec:data-mixes}.
To evaluate our fine-tuned model, we use test sets from two personalized segmentation benchmarks proposed by PDM~\cite{pdm} and PerSeg~\cite{perseg}.
PDM is curated from multiple datasets while PerSeg is synthesized from generative models.
Furthermore, we also evaluate our model on the test set from LASOT~\cite{lasot} dataset, where we divide the categories present in these datasets in equal proportions, \ie~$50\%$ of the total categories used for training and the remaining are used for testing. 
This is to ensure that the true generalization abilities of the model are evaluated and the distribution of train and test data is different. 
For the two segmentation benchmarks -- PDM~\cite{pdm} and PerSeg~\cite{perseg} used for evaluating the personalized segmentation abilities of the model, 
we transform the per-pixel segmentation masks to bounding box coordinates. 
We delegate more details about these datasets to the appendix. 
In Table~\ref{tab:dataset-statistics} we list the number of unique categories, objects/image, and the size of the evaluation sets for all $3$ datasets.
Note that the objects/image is an \emph{average} of the number of objects present per image in the entire test set. 

\paragraph{Implementation Details:} We evaluate and fine-tune the models by using the open-source codebase of the Llama-Factory~\cite{zheng2024llamafactory}. 
The fine-tuning parameters for LoRA are also selected as the default parameters used in the same codebase. 
For the main results, we choose to fine-tune Qwen2-VL~\cite{qwen2vl}, however, we also ablate with fine-tuning other models and find that our data curation and fine-tuning strategy is not model specific.
To encourage reproducibility, our entire codebase is provided as part of the appendix and will be released upon acceptance.  
\begin{table}[t!]
\small
    \centering
    \resizebox{0.7\linewidth}{!}{\begin{tabular}{l|c|c|c}
    \toprule
        & \textbf{Categories} & \textbf{Objects/Image} & \textbf{Size} \\ \midrule\midrule
        PDM & 106 & 4.2 & 1952 \\ \midrule
        PerSeg & 20 & 1 & 154 \\ \midrule
        ICL-LASOT & 36 & 2.4 & 576  \\
        \bottomrule \bottomrule
    \end{tabular}}
        \tabvcap
    \caption{\textbf{Test set statistics.} We list the total number of categories in the evaluation set of PDM, PerSeg and the unique categories in the test set formed from LASOT~\cite{lasot} dataset. The objects/image is an average number of objects present in the image, and size denotes the total number of conversations in the test set.}
    \label{tab:dataset-statistics}
        \tabvcap
\end{table}

\begin{table*}[t!]
\scriptsize
    \centering
    \renewcommand{\arraystretch}{1.2}
    \resizebox{0.8\textwidth}{!}{
\begin{tabular}{l| c c| c c c c| c c c c| c}
    \toprule
    \multirow{2}{*}{\textbf{Model}} & \multicolumn{2}{c|}{\textbf{PDM}} & \multicolumn{4}{c|}{\textbf{PerSeg}} & \multicolumn{4}{c|}{\textbf{ICL-LASOT}} & \multirow{2}{*}{\textbf{Average}} \\
    & 1-shot & 2-shot & 1-shot & 2-shot & 3-shot & 4-shot & 1-shot & 2-shot & 4-shot & 8-shot & \\
    \midrule
    \midrule
    Idefics3 & \phantom{0}4.14 & \phantom{0}3.76 & 12.77 & 16.22 & 19.54 & 12.32 & \phantom{0}3.32 & \phantom{0}3.94 & \phantom{0}8.21 & \phantom{0}0.01 & \phantom{0}8.42 \\
    Pixtral-12B & \phantom{0}3.63 & \phantom{0}4.42 & 20.04 & 21.44 & 37.31 & \underline{30.33} & \phantom{0}4.23 & \phantom{0}4.19 & \phantom{0}8.51 & 11.20 & 14.53 \\
    \llava-OV & 11.10 & 13.85 & \underline{43.01} & \underline{40.08} & 30.01 & 14.03 & 12.45 & 15.66 & 18.64 & \phantom{0}7.62 & 20.64 \\
    Qwen2-VL-7B & 16.75 & 19.72 & 34.21 & 30.22 & \underline{35.09} & 18.16 & 46.69 & \underline{49.46} & \underline{46.29} & \underline{21.12} & \underline{31.77} \\
    \midrule
        \method (7B) (Real) & \textbf{31.54} & \underline{24.45} & \textbf{43.40} & 29.90 & 33.23 & 25.25 & \underline{46.98} & 42.50 & 16.99 & 15.52 & 30.98 \\
    \method (7B) (Real + Pseudo) & \underline{29.21} & \textbf{25.88} & 42.43 & \textbf{41.15} & \textbf{39.85} & \textbf{33.08} & \textbf{49.71} & \textbf{57.14} & \textbf{59.41} & \textbf{55.21} & \textbf{43.31} \\
    \midrule
    \midrule
    GPT-4o & 11.27 & 13.29 & 33.43 & 41.41 & 44.77 & \underline{45.72} & 18.13 & 20.61 & 23.08 & 24.61 & 27.63 \\
    Qwen2-VL-72B & 16.88 & 12.97 & 41.88 & 35.47 & 49.51 & 39.96 & 42.66 & 38.05 & 33.54 & 31.17 & 34.21 \\
    InternVL2-76B & 31.26 & 28.67 & 50.01 & \underline{51.46} & 49.37 & 45.70 & 47.68 & 42.90 & 38.90 & 27.21 & 41.32 \\
    \midrule
    \method (72B) (Real) & \underline{42.30} & \underline{37.80} & \underline{50.21} & 50.99 & \underline{51.21} & 41.75 & \underline{59.89} & \underline{56.75} & \underline{54.57} & \underline{34.30} & \underline{47.98} \\
    \method (72B) (Real + Pseudo) & \textbf{43.69} & \textbf{46.60} & \textbf{52.32} & \textbf{53.21} & \textbf{53.11} & \textbf{47.82} & \textbf{64.71} & \textbf{65.71} & \textbf{67.63} & \textbf{63.15} & \textbf{55.79} \\
    \bottomrule
    \bottomrule
\end{tabular}}
    \tabvcap
    \caption{\textbf{Few-shot personalized object localization results.} We report the mIoU (\%) for different shots created from the test splits of the datasets we evaluate in this work. We evaluate up to the maximum number of shots we can create due to the variable number of samples present in these datasets. We obtain all our results by fine-tuning Qwen2-VL-7B and Qwen2-VL-72B~\cite{qwen2vl} on conversations consisting of only the \emph{real} names and also by the enhanced data with the \emph{pseudo} name regularization.}
    \tabvcap
    \label{tab:main-results}
\end{table*}

%
%
%


\subsection{Results}
\label{subsec:results}
For the main results, we evaluate GPT-4o~\cite{openai2023gpt4}, Idefics3~\cite{idefics3}, \llava-OV~\cite{llava-ov}, Pixtral (12B)~\cite{pixtral}, InternVL2 (76B)~\cite{internvl2}, and Qwen2-VL (7B/72B)~\cite{qwen2vl}.
These models are recent state-of-the-art models in the ever-evolving VLM landscape. 
We also evaluate two fine-tuned versions of Qwen2-VL (7B/72B) with separate data mixes: one only uses the \emph{real} category names and the other version employs \emph{real} and \emph{pseudo} category names mix (\cf~Section~\ref{subsec:data-mixes}).

In Table~\ref{tab:main-results} we list the detailed results on the $3$ datasets we evaluate our fine-tuned Qwen2-VL\footnote{Our fine-tuning also generalizes across different VLMs (\cf Section~\ref{subsec:ablations}).}.
We observe that most of the state-of-the-art (including proprietary GPT-4o) VLMs show weak performance on the task of few-shot personalized localization. 
For example, from Table~\ref{tab:main-results}, we observe that while comparing the medium sized models (7B-12B -- upper half), the Idefics3 model is the weakest for the task of personalized localization, although it shows strong performance on general reasoning tasks. 
This shows that the generic instruction-tuning data, curated from publicly available datasets, can help this model for a variety of reasoning tasks, like VQA -- but does not focus on obtaining spatial understanding to produce structured outputs or force these models to focus on \emph{context}. 
These results also highlight the deficiency of these models in understanding fine-grained concepts related to reasoning about relations and attributes in an image, as discussed in~\cite{huang2024conme}. 
Similarly, we find that \llava-OV comes in second, by showing an average improvement of $10.75\%$ over the Idefics3~\cite{idefics3} model.
This can be because \llava-OV includes localization data and interleaved image-text data in its fine-tuning corpus.
However, this data might not be semantically coherent or sampled from sources which can make the model focus on a single object of interest, thus failing to instill a notion of \emph{personalization} in these models.
In Table~\ref{tab:main-results} we also find that Qwen2-VL remains competitive with our~\method and outperforms Idefics3 and \llava-OV by $21.82\%$ and $11.07\%$ on average. 
However, it still lags behind our \method with a fair margin. 
For example, for the PDM test set our best model (trained only with \emph{real} category names) outperforms the base model for both $1$ and $2$ shots while on average being $9.76\%$ better. 
Similarly, we find that our fine-tuned model also outperforms the base model for the PerSeg and LASOT test sets with $12.58\%$ average improvement. 

\begin{table}[t!]
    \centering
    \resizebox{0.7\linewidth}{!}{\begin{tabular}{c c c |c c  c c c}
        \toprule
        \textbf{LASOT} & \textbf{GOT} & \textbf{TAO} & \textbf{PDM} & \textbf{PerSeg} & \textbf{ICL-LASOT}\\
        \midrule
        \midrule
        \checkmark &&& 16.58&35.67&30.30\\
        & \checkmark && 18.46 &27.34&28.25\\
        &&\checkmark & 12.05 &28.25&14.06\\
        &\checkmark&\checkmark & 21.40 &36.90 & 27.69\\
        \checkmark&&\checkmark& 21.63&37.67  & 28.04 \\
        \checkmark&\checkmark&&  21.72& 34.93  & 28.86\\
        \midrule
        \checkmark &\checkmark & \checkmark & \textbf{ 27.55} & \textbf{39.13} & \textbf{55.36}\\ 
        \bottomrule
        \bottomrule
    \end{tabular}}
        \tabvcap
    \caption{\textbf{Ablating data sources.} We report the mIoU (\%) by fine-tuning the Qwen2-VL-7B~\cite{qwen2vl} on different data sources.  }
    \label{tab:abl:data-mixes}
        \tabvcap
\end{table}


\setlength{\textfloatsep}{5mm} 
From Table~\ref{tab:main-results} (lower-half) we also observe that increasing the model size generally has a positive effect on the ICL abilities, consistent with the literature in NLP~\cite{wei2023larger}. 
However, surprisingly, the proprietary GPT-4o model performs underwhelmingly for this task, exposing unique weakness in an otherwise strong model. 
During our evaluations we observed that GPT-4o constantly provided reasoning about the image, instead of providing a bounding box for localizing the object. 
We delegate the responses and our efforts with multiple variants of the prompting techniques used for fair comparison, to the appendix.
Overall, the results highlight that our IPLoc can generalize to different model sizes, obtaining an average gain of $14.47\%$, even while comparing with a larger InternVL2 backbone.\\
We also observe, that enhancing the training data with \emph{pseudo} name regularization also improves results.
For example, we observe an average improvement of $12.33\%$ and $7.81\%$  while comparing the model trained only with the \emph{real} category names for the 7B and 72B variant, respectively. 
These results highlight the effectiveness of the regularization as it makes the models focus more on the contextual cues, rather than simply relying on the pre-trained knowledge to localize the object in the query image. 


\begin{table}[t!]
    \centering
    \begin{minipage}{0.48\linewidth}
        \centering
        \resizebox{\linewidth}{!}{%
        \begin{tabular}{lcc}
            \toprule
            Dataset & Qwen2-VL (72B) & IPLoc \\
            \midrule\midrule
            GQA  & 64.66 & 63.37\\
            SEED & 77.64 & 77.66\\
            POPE & 88.70 & 87.33\\
            \bottomrule\bottomrule
        \end{tabular}}
            \tabvcap
        \caption{\textbf{Retention of
Generalization.} Ablation highlighting the retention of generalization abilities of our fine-tuned model for the task of interest.}
        \label{tab:retention_generalization}
    \end{minipage}
    \hfill
    \begin{minipage}{0.48\linewidth}
        \centering
        \resizebox{0.90\linewidth}{!}{%
        \begin{tabular}{lcc}
            \toprule
            Dataset & Random & \method \\
            \midrule\midrule
            PDM & 28.94 & 50.05 \\
            PerSeg & 17.55 & 27.55 \\
            ICL-LASOT & 24.14 & 39.13 \\
            \bottomrule
            \bottomrule
        \end{tabular}}
            \tabvcap
        \caption{\textbf{Semantically
coherent data.} Ablating the effect of sampling conversations from the same video (\method) vs. arbitrary video sequences.}
        \label{tab:semantically_coherent_data}
    \end{minipage}    \tabvcap
\end{table}

\subsection{Ablations}
\label{subsec:ablations}
In this section, we ablate the key aspects of our proposed approach. 
Specifically, first, we examine the contribution of each data source. 
Then we ablate the retention of generalization abilities after fine-tuning, the effect of semantically coherent data (focus of this work), the performance on data containing only~\emph{pseudo} names,  and finally conclude with fine-tuning other models beyond Qwen2-VL, a more fine-grained analysis of increasing the number of shots, and comparisons with vanilla few-shot object detectors.

\noindent\textbf{Contribution of Data Sources:} In Table~\ref{tab:abl:data-mixes} we ablate the contribution from different types of data sources employed for fine-tuning in this work. 
We find that the best performance is obtained by mixing data from all three datasets~\cite{tao, got, lasot}, highlighting the effectiveness of the diversity of data and our proposed data mix.

\noindent\textbf{Retention of Generalization:} It is imperative that models must not \textit{over-specialize} upon the proposed task (in-context localization) and retain the generalization abilities to also be able to perform strongly on a broader range of tasks. 
In Table~\ref{tab:retention_generalization} we compare our fine-tuned model with the base Qwen2-VL-7B (without fine-tuning) on common VL benchmarks like GQA~\cite{gqa}, SEED~\cite{fang2021seed}, and POPE~\cite{pope}. 
We find that our parameter-efficient approach restricts the model parameter update and shows a degradation of less than $1\%$ (on average), thus, highlighting the retention of generalization abilities in VLMs after fine-tuning. 

\noindent\textbf{Semantically Coherent Data:} In Table~\ref{tab:semantically_coherent_data} we compare fine-tuning with our proposed data curation scheme of sampling few-shot data from the~\emph{same} video sequence~\emph{vs.} creating few-shot instruction-tuning conversations from different video sequences. 
We find that our proposed methodology fares well. 
This highlights the importance of semantically coherent context for fine-tuning. 

\begin{table}[t!]
    \centering
    \tiny
    \begin{minipage}{0.48\linewidth}
        \centering
        \resizebox{0.95\linewidth}{!}{%
        \begin{tabular}{lcc}
            \toprule
            Model & Qwen2-VL (7B) & IPLoc \\
            \midrule \midrule
            mIoU (\%) & 16.55 & 44.76 \\
            \bottomrule \bottomrule
        \end{tabular}}
        \subcaption{\textbf{Focus on Context.}}
        \label{tab:fake-names-data}
    \end{minipage}
    \hfill
    \begin{minipage}{0.48\linewidth}
        \centering
        \resizebox{0.95\linewidth}{!}{%
        \begin{tabular}{lcc}
            \toprule
            Model & LLaVA-OV (7B) & IPLoc \\
            \midrule\midrule
            mIoU (\%) & 20.65 & 27.25 \\
            \bottomrule \bottomrule
        \end{tabular}}
        \subcaption{\textbf{Generalization of Finetuning.}}
        \label{tab:generalization-of-finetuning}
    \end{minipage}
        \tabvcap
    \caption{(a) We test the ability of the model to focus on context instead of relying on pre-trained knowledge by switching the true category names with pseudo-names in the LASOT test set. (b) We evaluate the generalization of fine-tuning on LLaVA-OV.}
    \label{tab:ablation-results}
        \tabvcap
\end{table}

\noindent\textbf{Focus on Context:} To evaluate if our fine-tuned model focuses on context or simply relies on the knowledge about common categories learned during the large-scale pre-training, we design a test where we replace the original category names in the LASOT~\cite{lasot} test set with \emph{pseudo} names and report the results in Table~\ref{tab:fake-names-data}.
We find that our model fares better than the baseline, highlighting that our approach makes the models pay attention to the contextual cues rather than solely relying on the knowledge learned during pre-training to localize objects in the query image.

\noindent\textbf{Generalization of Finetuning:} 
To highlight the applicability of our fine-tuning methodology across different VLMs (beyond Qwen2-VL, used for main results in Table~\ref{tab:main-results}), we report the results obtained by fine-tuning \llava-OV~\cite{llava-ov} in Table~\ref{tab:generalization-of-finetuning}, where we obtain $\sim7\%$ improvement over the (base) pre-trained model. 
\begin{table}[t!]
\centering
\resizebox{0.6\linewidth}{!}{%
\begin{tabular}{l|cccc}
\toprule
 & 1 & 2 & 4 & 8 \\
\midrule\midrule
CD-ViTO~\cite{cdfsod} & 33.14 & 33.64 & 34.90 & 32.38 \\
TFA \cite{tfa} & 25.77 & 29.08 & 29.81 & 30.97\\
\midrule
IPLoc   & \textbf{64.71} & \textbf{65.71} & \textbf{67.63} & \textbf{63.15 }\\
\bottomrule\bottomrule
\end{tabular}%
}    \tabvcap
\caption{\textbf{Comparison with few-shot object detectors (FSOD).} We list the mIOU (\%) for vision-only FSOD and our IPLoc for the ICL-LASOT dataset.}
\label{tab:non_vlm_fsod}
    \tabvcap
\end{table}

\begin{figure}[t!]
    \centering
        \vspace{-10pt}
    \includegraphics[width=0.46\textwidth]{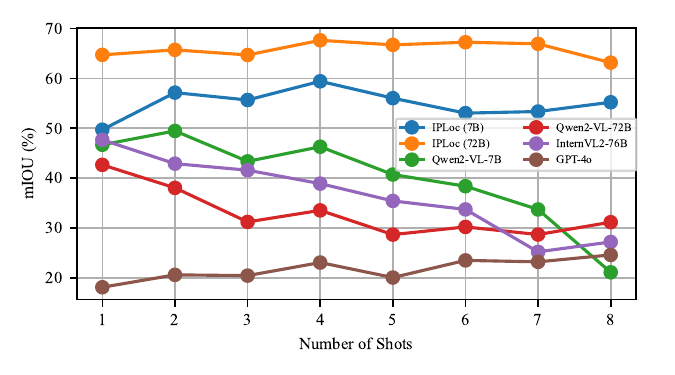} 
    \figvcap
    \caption{\textbf{Effect of increasing number of shots.} We report mIOU (\%) on the LASOT~\cite{lasot} test split. IPLoc refers to the Qwen2-VL~\cite{qwen2vl} fine-tuned on the proposed data mix in this paper.}
    \label{fig:shots_effect}
\end{figure}

\noindent\textbf{Increasing number of shots:} 
We provide a more fine-grained analysis of the effect of increasing number of shots on model performance in Figure~\ref{fig:shots_effect}. 
We observe that with increasing number of shots, the performance of most models decreases (potentially due to the inability of models to handle longer contexts~\cite{press2021train,ben2024decimamba} and generally lacking ICL abilities). 
In contrast, our IPLoc shows strong stability over multiple shots and across different model sizes. 
We note that the base Qwen2-VL (72B) performs worse than the 7B version. In the appendix, we detail the output responses and prompting techniques used (for a fair comparison). 
Further, we also observe that our proposed \method takes steps towards mitigating these issues, thanks to our tailored ICL instruction-tuning conversations.

\noindent\textbf{Cross-model Comparison:} In Tables~\ref{tab:non_vlm_fsod}~\&~\ref{tab:abl:ovd} we provide results by comparing with \textit{classic} vision-only few-shot object detectors~\cite{cdfsod,tfa}, and open-vocabulary object detector~\cite{owlvit}. 
We find that our IPLoc fares well on the task of personalized localization in cross-model comparisons. 

\begin{table}[t!]
    \centering
    \small
    \resizebox{0.6\linewidth}{!}{\begin{tabular}{l |c c c |c }
        \toprule
         & \textbf{PDM} & \textbf{PerSeg} & \textbf{ICL-LASOT} & Mean \\ 
         \midrule\midrule
        OwlViT~\cite{owlvit}    & \textbf{48.47} & 40.71 & 50.18 & 46.45 \\
        IPLoc              & 45.15 & \textbf{51.62} & \textbf{65.30} & \textbf{54.02} \\
        \bottomrule\bottomrule
        \end{tabular}}    \tabvcap

        \caption{\textbf{Comparison with open vocabulary object detector (OVD).} We list the mIOU (\%) for OVD and our \method.}
    \label{tab:abl:ovd}
        \tabvcap
\end{table}

\section{Limitations and Conclusion}
\label{sec:limitations}
\paragraph{Limitations:} As with any other research, our work also has certain known limitations. 
Currently, our setting involves localizing a single object of interest in the query examples, and we do not study localizing multiple objects in a single image. 
This is because current state-of-the-art VLMs struggle greatly when tasked with localizing multiple objects.
For example, Qwen2-VL scores 5.3$\%$ mIOU, when evaluated for the multi-object localization task in MS-COCO, which can be attributed to the pre-training data of Qwen2-VL, which potentially only contains conversations that require localization of single objects in images. 
In the future, as VLMs capable of localizing multiple objects become available, our \method can directly benefit. 

\paragraph{Conclusion:} We propose a data-centric approach to enhance the few-shot personalized object localization capabilities of modern VLMs. 
Our findings reveal that for VLMs to excel in this task, fine-tuning data must be semantically coherent, personalized, and designed to encourage reliance on contextual cues rather than pre-existing pre-training knowledge. 
To achieve this, we utilize multiple video object-tracking datasets and introduce innovative regularization schemes for constructing instruction-tuning conversations that provoke robust few-shot learning abilities in VLMs.
Through extensive evaluations and ablation studies across multiple benchmarks, we demonstrate the effectiveness of our proposed data and fine-tuning strategies. 
The results provide a fresh perspective on the impact of data-centric methods for advancing personalized localization capabilities in VLMs, offering valuable insights and a foundation for future work in this direction.

{\small
\bibliographystyle{ieee_fullname}
\bibliography{egbib}

\begin{thebibliography}{10}\itemsep=-1pt

\bibitem{pixtral}
Pravesh Agrawal, Szymon Antoniak, Emma~Bou Hanna, Baptiste Bout, Devendra Chaplot, Jessica Chudnovsky, Diogo Costa, Baudouin De~Monicault, Saurabh Garg, Theophile Gervet, et~al.
\newblock Pixtral 12b.
\newblock {\em arXiv preprint arXiv:2410.07073}, 2024.

\bibitem{flamingo}
Jean-Baptiste Alayrac, Jeff Donahue, Pauline Luc, Antoine Miech, Iain Barr, Yana Hasson, Karel Lenc, Arthur Mensch, Katie Millican, Malcolm Reynolds, Roman Ring, Eliza Rutherford, Serkan Cabi, Tengda Han, Zhitao Gong, Sina Samangooei, Marianne Monteiro, Jacob Menick, Sebastian Borgeaud, Andrew Brock, Aida Nematzadeh, Sahand Sharifzadeh, Mikolaj Binkowski, Ricardo Barreira, Oriol Vinyals, Andrew Zisserman, and Karen Simonyan.
\newblock {Flamingo: a Visual Language Model for Few-Shot Learning}.
\newblock In {\em NeurIPS}, 2022.

\bibitem{qwen-vl}
Jinze Bai, Shuai Bai, Shusheng Yang, Shijie Wang, Sinan Tan, Peng Wang, Junyang Lin, Chang Zhou, and Jingren Zhou.
\newblock Qwen-vl: A versatile vision-language model for understanding, localization, text reading, and beyond.
\newblock {\em arXiv preprint arXiv:2308.12966}, 2023.

\bibitem{ben2024decimamba}
Assaf Ben-Kish, Itamar Zimerman, Shady Abu-Hussein, Nadav Cohen, Amir Globerson, Lior Wolf, and Raja Giryes.
\newblock Decimamba: Exploring the length extrapolation potential of mamba.
\newblock {\em arXiv preprint arXiv:2406.14528}, 2024.

\bibitem{brown2020language}
Tom Brown, Benjamin Mann, Nick Ryder, Melanie Subbiah, Jared~D Kaplan, Prafulla Dhariwal, Arvind Neelakantan, Pranav Shyam, Girish Sastry, Amanda Askell, et~al.
\newblock Language models are few-shot learners.
\newblock {\em Advances in neural information processing systems}, 33:1877--1901, 2020.

\bibitem{gpt3}
Tom~B. Brown, Benjamin Mann, Nick Ryder, Melanie Subbiah, Jared Kaplan, Prafulla Dhariwal, Arvind Neelakantan, Pranav Shyam, Girish Sastry, Amanda Askell, Sandhini Agarwal, Ariel Herbert-Voss, Gretchen Krueger, Tom Henighan, Rewon Child, Aditya Ramesh, Daniel~M. Ziegler, Jeffrey Wu, Clemens Winter, Christopher Hesse, Mark Chen, Eric Sigler, Mateusz Litwin, Scott Gray, Benjamin Chess, Jack Clark, Christopher Berner, Sam McCandlish, Alec Radford, Ilya Sutskever, and Dario Amodei.
\newblock {Language Models are Few-Shot Learners}.
\newblock In {\em NeurIPS}, 2020.

\bibitem{minigptv2}
Jun Chen, Deyao Zhu, Xiaoqian Shen, Xiang Li, Zechun Liu, Pengchuan Zhang, Raghuraman Krishnamoorthi, Vikas Chandra, Yunyang Xiong, and Mohamed Elhoseiny.
\newblock {MiniGPT-v2: Large Language Model as a Unified Interface for Vision-Language Multi-task Learning}.
\newblock In {\em Proc. ICLR}, 2024.

\bibitem{internvl2}
Zhe Chen, Weiyun Wang, Hao Tian, Shenglong Ye, Zhangwei Gao, Erfei Cui, Wenwen Tong, Kongzhi Hu, Jiapeng Luo, Zheng Ma, et~al.
\newblock How far are we to gpt-4v? closing the gap to commercial multimodal models with open-source suites.
\newblock {\em arXiv preprint arXiv:2404.16821}, 2024.

\bibitem{vicuna2023}
Wei-Lin Chiang, Zhuohan Li, Zi Lin, Ying Sheng, Zhanghao Wu, Hao Zhang, Lianmin Zheng, Siyuan Zhuang, Yonghao Zhuang, Joseph~E. Gonzalez, Ion Stoica, and Eric~P. Xing.
\newblock {Vicuna: An Open-Source Chatbot Impressing GPT-4 with 90\%* ChatGPT Quality}, 2023.

\bibitem{instructblip}
Wenliang Dai, Junnan Li, Dongxu Li, Anthony Tiong, Junqi Zhao, Weisheng Wang, Boyang Li, Pascale Fung, and Steven Hoi.
\newblock {InstructBLIP: Towards General-purpose Vision-Language Models with Instruction Tuning}.
\newblock In {\em NeurIPS}, 2023.

\bibitem{tao}
Achal Dave, Tarasha Khurana, Pavel Tokmakov, Cordelia Schmid, and Deva Ramanan.
\newblock Tao: A large-scale benchmark for tracking any object.
\newblock In {\em Computer Vision--ECCV 2020: 16th European Conference, Glasgow, UK, August 23--28, 2020, Proceedings, Part V 16}, pages 436--454. Springer, 2020.

\bibitem{dac}
Sivan Doveh, Assaf Arbelle, Sivan Harary, Amit Alfassy, Roei Herzig, Donghyun Kim, Raja Giryes, Rogerio Feris, Rameswar Panda, Shimon Ullman, et~al.
\newblock {Dense and Aligned Captions (DAC) Promote Compositional Reasoning in VL Models}.
\newblock In {\em NeurIPS}, 2023.

\bibitem{svlc}
Sivan Doveh, Assaf Arbelle, Sivan Harary, Rameswar Panda, Roei Herzig, Eli Schwartz, Donghyun Kim, Raja Giryes, Rogerio Feris, Shimon Ullman, and Leonid Karlinsky.
\newblock Teaching structured vision \& language concepts to vision \& language models.
\newblock In {\em Proc. CVPR}, 2023.

\bibitem{llava_icl}
Sivan Doveh, Shaked Perek, M~Jehanzeb Mirza, Amit Alfassy, Assaf Arbelle, Shimon Ullman, and Leonid Karlinsky.
\newblock {Towards multimodal in-context learning for vision \& language models}.
\newblock {\em arXiv preprint arXiv:2403.12736}, 2024.

\bibitem{dubey2024llama}
Abhimanyu Dubey, Abhinav Jauhri, Abhinav Pandey, Abhishek Kadian, Ahmad Al-Dahle, Aiesha Letman, Akhil Mathur, Alan Schelten, Amy Yang, Angela Fan, et~al.
\newblock {The Llama 3 Herd of Models}.
\newblock {\em arXiv preprint arXiv:2407.21783}, 2024.

\bibitem{lasot}
Heng Fan, Liting Lin, Fan Yang, Peng Chu, Ge Deng, Sijia Yu, Hexin Bai, Yong Xu, Chunyuan Liao, and Haibin Ling.
\newblock Lasot: A high-quality benchmark for large-scale single object tracking.
\newblock In {\em Proceedings of the IEEE/CVF conference on computer vision and pattern recognition}, pages 5374--5383, 2019.

\bibitem{fang2021seed}
Zhiyuan Fang, Jianfeng Wang, Lijuan Wang, Lei Zhang, Yezhou Yang, and Zicheng Liu.
\newblock {SEED: Self-supervised Distillation for Visual Representation}.
\newblock In {\em Proc. ICLR}, 2021.

\bibitem{cdfsod}
Yuqian Fu, Yu Wang, Yixuan Pan, Lian Huai, Xingyu Qiu, Zeyu Shangguan, Tong Liu, Yanwei Fu, Luc Van~Gool, and Xingqun Jiang.
\newblock Cross-domain few-shot object detection via enhanced open-set object detector.
\newblock In {\em European Conference on Computer Vision}, pages 247--264. Springer, 2025.

\bibitem{tv_mm}
Amir~Bar Grace~Luo, Trevor~Darrell.
\newblock Task vectors are cross-modal.
\newblock {\em arXiv preprint arXiv:2410.22330}, 2024.

\bibitem{lora}
Edward~J Hu, Yelong Shen, Phillip Wallis, Zeyuan Allen-Zhu, Yuanzhi Li, Shean Wang, Lu Wang, and Weizhu Chen.
\newblock Lora: Low-rank adaptation of large language models.
\newblock {\em arXiv preprint arXiv:2106.09685}, 2021.

\bibitem{task-vec}
Brandon Huang, Chancharik Mitra, Assaf Arbelle, Leonid Karlinsky, Trevor Darrell, and Roei Herzig.
\newblock Multimodal task vectors enable many-shot multimodal in-context learning.
\newblock {\em arXiv preprint arXiv:2406.15334}, 2024.

\bibitem{huang2024conme}
Irene Huang, Wei Lin, M~Jehanzeb Mirza, Jacob~A Hansen, Sivan Doveh, Victor~Ion Butoi, Roei Herzig, Assaf Arbelle, Hilde Kuhene, Trevor Darrel, et~al.
\newblock {ConMe: Rethinking Evaluation of Compositional Reasoning for Modern VLMs}.
\newblock {\em arXiv preprint arXiv:2406.08164}, 2024.

\bibitem{got}
Lianghua Huang, Xin Zhao, and Kaiqi Huang.
\newblock Got-10k: A large high-diversity benchmark for generic object tracking in the wild.
\newblock {\em IEEE transactions on pattern analysis and machine intelligence}, 43(5):1562--1577, 2019.

\bibitem{gqa}
Drew~A Hudson and Christopher~D Manning.
\newblock Gqa: A new dataset for real-world visual reasoning and compositional question answering.
\newblock In {\em Proceedings of the IEEE/CVF conference on computer vision and pattern recognition}, pages 6700--6709, 2019.

\bibitem{align}
Chao Jia, Yinfei Yang, Ye Xia, Yi-Ting Chen, Zarana Parekh, Hieu Pham, Quoc~V. Le, Yunhsuan Sung, Zhen Li, and Tom Duerig.
\newblock {Scaling Up Visual and Vision-Language Representation Learning With Noisy Text Supervision}.
\newblock In {\em Proc. ICML}, 2021.

\bibitem{clippr}
Jonathan Kahana, Niv Cohen, and Yedid Hoshen.
\newblock {Improving Zero-Shot Models with Label Distribution Priors}.
\newblock {\em arXiv:2212.00784}, 2022.

\bibitem{idefics3}
Hugo Laurençon, Andrés Marafioti, Victor Sanh, and Léo Tronchon.
\newblock Building and better understanding vision-language models: insights and future directions., 2024.

\bibitem{idefics}
Hugo Laurençon, Lucile Saulnier, Léo Tronchon, Stas Bekman, Amanpreet Singh, Anton Lozhkov, Thomas Wang, Siddharth Karamcheti, Alexander~M. Rush, Douwe Kiela, Matthieu Cord, and Victor Sanh.
\newblock Obelics: An open web-scale filtered dataset of interleaved image-text documents, 2023.

\bibitem{llava-ov}
Bo Li, Yuanhan Zhang, Dong Guo, Renrui Zhang, Feng Li, Hao Zhang, Kaichen Zhang, Yanwei Li, Ziwei Liu, and Chunyuan Li.
\newblock {LLaVA-OneVision: Easy Visual Task Transfer}.
\newblock {\em arXiv preprint arXiv:2408.03326}, 2024.

\bibitem{blip2}
Junnan Li, Dongxu Li, Silvio Savarese, and Steven Hoi.
\newblock {BLIP-2: Bootstrapping Language-Image Pre-training with Frozen Image Encoders and Large Language Models}.
\newblock In {\em Proc. ICML}, 2023.

\bibitem{pope}
Yifan Li, Yifan Du, Kun Zhou, Jinpeng Wang, Wayne~Xin Zhao, and Ji-Rong Wen.
\newblock Evaluating object hallucination in large vision-language models.
\newblock {\em arXiv preprint arXiv:2305.10355}, 2023.

\bibitem{lin2023video}
Bin Lin, Yang Ye, Bin Zhu, Jiaxi Cui, Munan Ning, Peng Jin, and Li Yuan.
\newblock Video-{LL}a{VA}: Learning united visual representation by alignment before projection.
\newblock In {\em Proceedings of the Conference on Empirical Methods in Natural Language Processing (EMNLP)}, pages 5971--5984, 2024.

\bibitem{mscoco}
Tsung-Yi Lin, Michael Maire, Serge Belongie, James Hays, Pietro Perona, Deva Ramanan, Piotr Doll{\'a}r, and C~Lawrence Zitnick.
\newblock Microsoft coco: Common objects in context.
\newblock In {\em Computer Vision--ECCV 2014: 13th European Conference, Zurich, Switzerland, September 6-12, 2014, Proceedings, Part V 13}, pages 740--755. Springer, 2014.

\bibitem{maxi}
Wei Lin, Leonid Karlinsky, Nina Shvetsova, Horst Possegger, Mateusz Kozinski, Rameswar Panda, Rogerio Feris, Hilde Kuehne, and Horst Bischof.
\newblock {MAtch, eXpand and Improve: Unsupervised Finetuning for Zero-Shot Action Recognition with Language Knowledge}.
\newblock In {\em Proc. ICCV}, 2023.

\bibitem{llava-next}
Haotian Liu, Chunyuan Li, Yuheng Li, and Yong~Jae Lee.
\newblock {LLaVA-NeXT: Improved reasoning, OCR, and world knowledge}, 2023.

\bibitem{llava+}
Haotian Liu, Chunyuan Li, Yuheng Li, and Yong~Jae Lee.
\newblock {Improved Baselines with Visual Instruction Tuning}.
\newblock In {\em Proc. CVPR}, 2024.

\bibitem{llava}
Haotian Liu, Chunyuan Li, Qingyang Wu, and Yong~Jae Lee.
\newblock {Visual Instruction Tuning}.
\newblock In {\em NeurIPS}, 2023.

\bibitem{icl}
Sewon Min, Mike Lewis, Luke Zettlemoyer, and Hannaneh Hajishirzi.
\newblock {MetaICL: Learning to Learn In Context}.
\newblock In {\em Proc. NAACL}, 2022.

\bibitem{owlvit}
Matthias Minderer, Alexey Gritsenko, Austin Stone, Maxim Neumann, Dirk Weissenborn, Alexey Dosovitskiy, Aravindh Mahendran, Anurag Arnab, Mostafa Dehghani, Zhuoran Shen, et~al.
\newblock Simple open-vocabulary object detection.
\newblock In {\em European conference on computer vision}, pages 728--755. Springer, 2022.

\bibitem{mpvr}
M.~Jehanzeb Mirza, Leonid Karlinsky, Wei Lin, Sivan Doveh, , Jakub Micorek, Mateusz Kozinski, Hilde Kuhene, and Horst Possegger.
\newblock {Meta-Prompting for Automating Zero-shot Visual Recognition with LLMs}.
\newblock In {\em Proc. ECCV}, 2024.

\bibitem{tap}
M.~Jehanzeb Mirza, Leonid Karlinsky, Wei Lin, Horst Possegger, Rogerio Feris, and Horst Bischof.
\newblock {TAP: Targeted Prompting for Task Adaptive Generation of Textual Training Instances for Visual Classification}.
\newblock {\em arXiv preprint arXiv:2309.06809}, 2023.

\bibitem{lafter}
Muhammad~Jehanzeb Mirza, Leonid Karlinsky, Wei Lin, Horst Possegger, Mateusz Kozinski, Rogerio Feris, and Horst Bischof.
\newblock {La{FT}er: Label-Free Tuning of Zero-shot Classifier using Language and Unlabeled Image Collections}.
\newblock In {\em NeurIPS}, 2023.

\bibitem{glov}
M~Jehanzeb Mirza, Mengjie Zhao, Zhuoyuan Mao, Sivan Doveh, Wei Lin, Paul Gavrikov, Michael Dorkenwald, Shiqi Yang, Saurav Jha, Hiromi Wakaki, et~al.
\newblock Glov: Guided large language models as implicit optimizers for vision language models.
\newblock {\em arXiv preprint arXiv:2410.06154}, 2024.

\bibitem{openai2023gpt4}
OpenAI.
\newblock {GPT-4 Technical Report}.
\newblock {\em arXiv preprint arXiv:2303.08774}, 2023.

\bibitem{press2021train}
Ofir Press, Noah~A Smith, and Mike Lewis.
\newblock Train short, test long: Attention with linear biases enables input length extrapolation.
\newblock {\em arXiv preprint arXiv:2108.12409}, 2021.

\bibitem{clip}
Alec Radford, Jong~Wook Kim, Chris Hallacy, Aditya Ramesh, Gabriel Goh, Sandhini Agarwal, Girish Sastry, Amanda Askell, Pamela Mishkin, Jack Clark, Gretchen Krueger, and Ilya Sutskever.
\newblock {Learning Transferable Visual Models from Natural Language Supervision}.
\newblock In {\em Proc. ICML}, 2021.

\bibitem{pdm}
Dvir Samuel, Rami Ben-Ari, Matan Levy, Nir Darshan, and Gal Chechik.
\newblock Where's waldo: Diffusion features for personalized segmentation and retrieval.
\newblock {\em NeurIPS}, 2024.

\bibitem{openclip}
Christoph Schuhmann, Romain Beaumont, Richard Vencu, Cade~W Gordon, Ross Wightman, Mehdi Cherti, Theo Coombes, Aarush Katta, Clayton Mullis, Mitchell Wortsman, Patrick Schramowski, Srivatsa~R Kundurthy, Katherine Crowson, Ludwig Schmidt, Robert Kaczmarczyk, and Jenia Jitsev.
\newblock {LAION}-5b: An open large-scale dataset for training next generation image-text models.
\newblock In {\em NeurIPS}, 2022.

\bibitem{emu2}
Quan Sun, Yufeng Cui, Xiaosong Zhang, Fan Zhang, Qiying Yu, Zhengxiong Luo, Yueze Wang, Yongming Rao, Jingjing Liu, Tiejun Huang, et~al.
\newblock Generative multimodal models are in-context learners.
\newblock {\em arXiv preprint arXiv:2312.13286}, 2023.

\bibitem{qwen2vl}
Peng Wang, Shuai Bai, Sinan Tan, Shijie Wang, Zhihao Fan, Jinze Bai, Keqin Chen, Xuejing Liu, Jialin Wang, Wenbin Ge, et~al.
\newblock Qwen2-vl: Enhancing vision-language model's perception of the world at any resolution.
\newblock {\em arXiv preprint arXiv:2409.12191}, 2024.

\bibitem{tfa}
Xin Wang, Thomas~E Huang, Trevor Darrell, Joseph~E Gonzalez, and Fisher Yu.
\newblock Frustratingly simple few-shot object detection.
\newblock {\em arXiv preprint arXiv:2003.06957}, 2020.

\bibitem{cot}
Jason Wei, Xuezhi Wang, Dale Schuurmans, Maarten Bosma, Fei Xia, Ed Chi, Quoc~V Le, Denny Zhou, et~al.
\newblock {Chain-of-Thought Prompting Elicits Reasoning in Large Language Models}.
\newblock In {\em NeurIPS}, 2022.

\bibitem{wei2023larger}
Jerry Wei, Jason Wei, Yi Tay, Dustin Tran, Albert Webson, Yifeng Lu, Xinyun Chen, Hanxiao Liu, Da Huang, Denny Zhou, et~al.
\newblock Larger language models do in-context learning differently.
\newblock {\em arXiv preprint arXiv:2303.03846}, 2023.

\bibitem{metaclip}
Hu Xu, Saining Xie, Xiaoqing~Ellen Tan, Po-Yao Huang, Russell Howes, Vasu Sharma, Shang-Wen Li, Gargi Ghosh, Luke Zettlemoyer, and Christoph Feichtenhofer.
\newblock {Demystifying CLIP Data}.
\newblock In {\em Proc. ICLR}, 2023.

\bibitem{treeofthought}
Shunyu Yao, Dian Yu, Jeffrey Zhao, Izhak Shafran, Tom Griffiths, Yuan Cao, and Karthik Narasimhan.
\newblock {Tree of Thoughts: Deliberate Problem Solving with Large Language Models}.
\newblock In {\em NeurIPS}, 2023.

\bibitem{siglip}
Xiaohua Zhai, Basil Mustafa, Alexander Kolesnikov, and Lucas Beyer.
\newblock {Sigmoid Loss for Language Image Pre-training}.
\newblock In {\em Proc. ICCV}, 2023.

\bibitem{perseg}
Renrui Zhang, Zhengkai Jiang, Ziyu Guo, Shilin Yan, Junting Pan, Xianzheng Ma, Hao Dong, Peng Gao, and Hongsheng Li.
\newblock Personalize segment anything model with one shot.
\newblock {\em arXiv preprint arXiv:2305.03048}, 2023.

\bibitem{mmicl}
Haozhe Zhao, Zefan Cai, Shuzheng Si, Xiaojian Ma, Kaikai An, Liang Chen, Zixuan Liu, Sheng Wang, Wenjuan Han, and Baobao Chang.
\newblock Mmicl: Empowering vision-language model with multi-modal in-context learning.
\newblock {\em arXiv preprint arXiv:2309.07915}, 2023.

\bibitem{zheng2024llamafactory}
Yaowei Zheng, Richong Zhang, Junhao Zhang, Yanhan Ye, Zheyan Luo, Zhangchi Feng, and Yongqiang Ma.
\newblock Llamafactory: Unified efficient fine-tuning of 100+ language models.
\newblock In {\em Proceedings of the 62nd Annual Meeting of the Association for Computational Linguistics (Volume 3: System Demonstrations)}, Bangkok, Thailand, 2024. Association for Computational Linguistics.

\bibitem{minigpt}
Deyao Zhu, Jun Chen, Xiaoqian Shen, Xiang Li, and Mohamed Elhoseiny.
\newblock {MiniGPT-4: Enhancing Vision-Language Understanding with Advanced Large Language Models}.
\newblock In {\em Proc. ICLR}, 2024.

\end{thebibliography}
}
\clearpage
\appendix
\section*{Appendix}
\setcounter{section}{0}
In the following, we provide additional experiments and further explanations that offer deeper insights and enhance the clarity of the main manuscript.
In Section~\ref{sec:full_bars}, we present the detailed results for Qwen2-VL-7B retention of generalization, and the LLaVA-OV model (with and without) finetuning, complementing the ablation results from the main text.
In Section~\ref{sec:convert}, we detail the methodology for converting pixel-level segmentation maps from datasets like PDM and PerSeg into bounding box annotations, ensuring compatibility with our application.
In Section~\ref{sec:visualizations}, we provide extensive qualitative visualizations across various datasets and few-shot settings, highlighting both successful localizations and challenges in instance-level discrimination. 
Comprehensive examples for 1-shot to 8-shot scenarios are included to illustrate the robustness and adaptability of our method.
Finally, in Section~\ref{sec:prompts},
we include a prompt ablation study that explores how different prompt variants influence the model’s localization performance, thereby highlighting best practices for designing prompts in inference.

\section{Detailed Results}
\label{sec:full_bars}
In the following, we first provide extended results of experiments performed in the main manuscript highlighting the retention of generalization abilities of our fine-tuned models, then provide results showing the generalization of fine-tuning using our method across network architectures.

\subsection{Retention of generalization}
We extend Table 4 of the main paper for the 7B model.
In Table \ref{tab:small_model_retention_generalization} we show the minimal performance degradation of IPLoc-7B compared to the base model on several tasks, namely GQA, Seed-Bench and POPE.


\subsection{Retention of generalization across architectures}
The detailed results obtained from base LLaVA-OV and the fine-tuned model (on our dataset) are presented in Table \ref{tab:llava-ov}.
We find that our \method consistently improves the base model on all the few-shot splits we test on. 

These results provide insights regarding the generalization of our fine-tuning methodology across different vision language models (VLMs).
Note that in the main manuscript (Table 2), we fine-tuned Qwen2-VL~\cite{qwen2vl}.
\begin{table}
     \centering
        \resizebox{0.6\linewidth}{!}{%
        \begin{tabular}{lcc}
            \toprule
            \textbf{Dataset} & \textbf{Qwen2-VL-7B} & \textbf{IPLoc} \\
            \midrule\midrule
            GQA & 62.34 & 61.11 \\
            SEED & 72.06 & 71.16 \\
            POPE & 88.35 & 88.04 \\
            \bottomrule\bottomrule
        \end{tabular}}
        \caption{\textbf{Retention of
Generalization.} Ablation highlighting the retention of generalization abilities of our fine-tuned model for the task of interest.}
        \label{tab:small_model_retention_generalization}
\end{table}

\begin{table}[t!]
\centering
\begin{tabular}{llcc}
\toprule
\textbf{Dataset} & \textbf{Shots} & \textbf{LLaVA-OV} & \textbf{IPLoc-LLaVA-OV} \\
\midrule\midrule
PDM & 1 & 11.10 & 12.29 \\
& 2 & 13.85 & 15.03 \\
\midrule
PerSeg & 1 & 43.01 & 52.96 \\
& 2 & 40.08 & 57.90 \\
& 3 & 30.01 & 56.51 \\
& 4 & 14.03 & 18.11 \\
\midrule
ICL-LASOT& 1 & 12.45 & 13.99 \\
& 2 & 15.66 & 16.88 \\
& 4 & 18.64 & 21.04 \\
& 8 & \phantom{0}7.62 & \phantom{0}7.80 \\
\midrule
Average & & 20.65 & 27.25 \\
\bottomrule\bottomrule
\end{tabular}
\caption{\textbf{Generalization across different VLMs.} We report the detailed results of base and IPLoc-finetuned LLaVa-OV model.}
\label{tab:llava-ov}
\end{table}

\section{Segmentation Masks to Bounding Boxes} 
\label{sec:convert}
The PDM \cite{pdm} and PerSeg \cite{perseg} datasets provide images annotated with pixel-level segmentation maps, where each item is uniquely labeled. Since our method requires bounding box annotations, we directly utilize the segmentation maps.
Each segmentation map, containing unique object labels, is processed to extract bounding box coordinates $(x_1, y_1, x_2, y_2)$ that enclose the objects. These bounding box annotations are derived from the segmentation maps without the need for additional pixel-level computation, as the mapping between labels and objects is predefined.

\section{Comprehensive Visual Analysis Across Datasets}
\label{sec:visualizations}

We present an extensive set of qualitative results demonstrating our method's performance across multiple benchmark datasets (PDM, PerSeg, ICL-LASOT) under various few-shot settings. Our visualization framework employs a consistent color-coding scheme where red bounding boxes denote ground truth annotations in the support frames, while blue bounding boxes indicate our model's predictions on query frames. For each example, we show the support shots (containing the target object with its localization) followed by the corresponding query image in the final column.

\newcommand{\figref}[1]{Figure~\ref{#1}}

\begin{figure*}[t]
    \centering
    \includegraphics[width=0.95\textwidth,trim={0 0 0 0},clip]{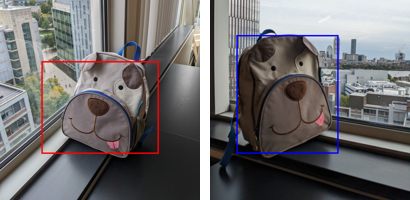}
    \caption{\textbf{One-shot Localization Results.} Representative examples of object localization using a single support frame in the PerSeg dataset.}
    \label{fig:1shot_success}
\end{figure*}

\begin{figure*}[t]
    \centering
    \includegraphics[width=0.95\textwidth,trim={0 0 0 0},clip]{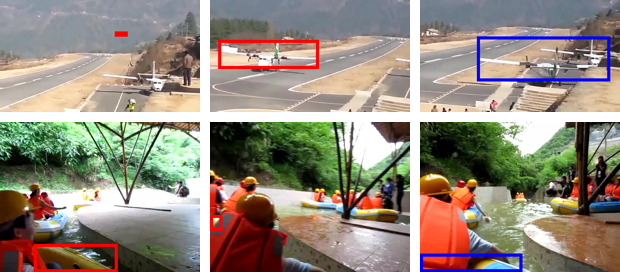}
    \caption{\textbf{Two-shot Localization Results.} Performance visualization with two support frames on the LaSOT and PDM datasets.}
    \label{fig:2shot}
\end{figure*}

\begin{figure*}[t]
    \centering
    \includegraphics[width=0.95\textwidth,trim={0 0 0 0},clip]{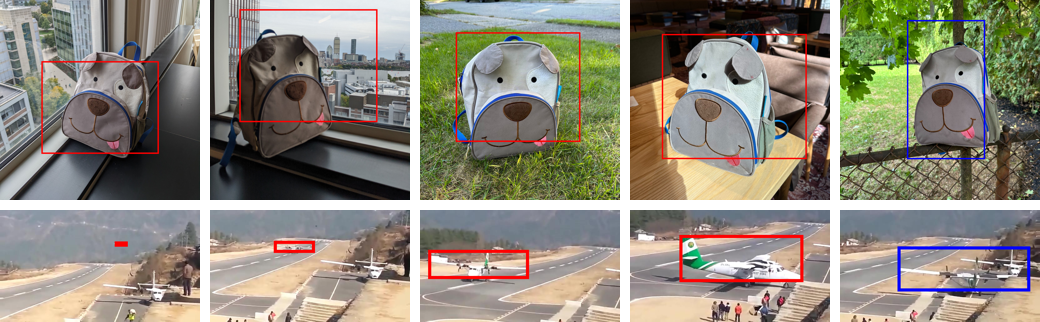}
    \caption{\textbf{Four-shot Localization Examples.} Visualization of our method's localization results when provided with four support frames on PerSeg and LaSOT datasets.}
    \label{fig:4shot}
\end{figure*}

\begin{figure*}[t]
    \centering
    \includegraphics[width=0.95\textwidth,trim={0 0 0 0},clip]{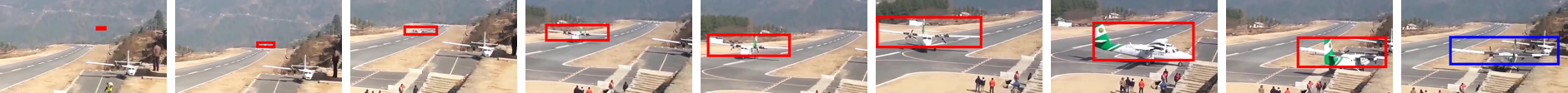}
    \caption{\textbf{Eight-shot Setting Results.} Demonstration of our method's localization behavior in the eight-shot setting on the LaSOT dataset.}
    \label{fig:8shot}
\end{figure*}

\begin{figure*}[t]
    \centering
    \includegraphics[width=0.95\textwidth,trim={0 0 0 0},clip]{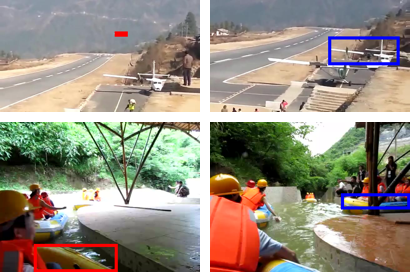}
    \caption{\textbf{Challenging Cases in One-shot Setting.} Examples where the model identifies semantically similar objects (incorrect airplane and boat) but fails to distinguish the specific target instance, highlighting the complexity of instance-level discrimination.}
    \label{fig:1shot_failure}
\end{figure*}

As shown in \figref{fig:1shot_success}, our method successfully localizes objects with just a single support frame across diverse scenarios. 
Further, in \figref{fig:2shot}, \figref{fig:4shot} and \figref{fig:8shot} we can see our method also performs well with more shots.
We also highlight some challenges of instance-level discrimination in \figref{fig:1shot_failure}, where semantically similar objects lead to incorrect localizations.

\section{Prompt Ablation}

\begin{table*}[t!]
\scriptsize
    \centering
    \renewcommand{\arraystretch}{1.2}
    \resizebox{1.\textwidth}{!}{
\begin{tabular}{l l| c c| c c c c| c c c c| c}
    \toprule
    \multirow{2}{*}{\textbf{Model}} & \multirow{2}{*}{\textbf{Prompt}} & \multicolumn{2}{c|}{\textbf{PDM}} & \multicolumn{4}{c|}{\textbf{PerSeg}} & \multicolumn{4}{c|}{\textbf{ICL-LASOT}} & \multirow{2}{*}{\textbf{Average}} \\
    & & 1-shot & 2-shot & 1-shot & 2-shot & 3-shot & 4-shot & 1-shot & 2-shot & 4-shot & 8-shot & \\
    \midrule
    \midrule
    Qwen2-VL-72B & Org. Prompt (~\ref{fig:org_prompt}) & 33.65 & 32.37 & 0.81 & 0.90 & 0.29 & 1.26 & 4.82 & 0.40 & 1.63 & 0.76 & 7.69 \\
    Qwen2-VL-72B & Prompt 1 (~\ref{fig:prompt1}) & 16.88 & 12.97 & 41.88 & 35.47 & 49.51 & 39.96 & 42.66 & 38.05 & 33.54 & 31.17 & \textbf{34.21} \\
    Qwen2-VL-72B & Prompt 2 (~\ref{fig:prompt2}) & 19.96 &13.11&44.75& 28.19& 43.43&37.36& 43.14&40.04&31.29&22.81&32.41 \\
    Qwen2-VL-72B & Prompt 3 (~\ref{fig:prompt3}) & 21.23 &12.29& 34.91& 28.68&45.50& 31.04&43.01 &37.65&31.38&27.23&31.29\\
    \midrule
    \midrule
    InternVL2-76B & Org. Prompt (~\ref{fig:org_prompt}) & 23.28& 14.20& 34.51&24.18 & 44.94& 36.57& 40.67 & 23.21 & 31.72 & 21.65 & 29.49\\
    InternVL2-76B & Prompt 1 (~\ref{fig:prompt1}) & 27.48 &21.87&49.01&41.16& 43.22&34.63 & 42.47& 45.61 & 39.71 & 21.74 & 36.69\\
    InternVL2-76B & Prompt 2 (~\ref{fig:prompt2}) & 31.26 & 28.67 & 50.01 & 51.46 & 49.37 & 45.70 & 47.68 & 42.90 & 38.90 & 27.21 & \textbf{41.32} \\
    InternVL2-76B & Prompt 3 (~\ref{fig:prompt3}) & 29.11 & 20.85 &46.06& 42.30& 50.03&42.34&46.07 & 43.76 & 38.86 & 29.18 & 38.86\\
    \bottomrule
    \bottomrule
\end{tabular}}
    \tabvcap
    \caption{\textbf{Prompt ablation for Qwen2-VL-72B and InternVL2-76B.} We employ multiple prompt strategies to gain the best performance (\% IoU) out of the big LMMs (Qwen2-VL-72B, InternVL2-76B). Suprisingly, the default prompt which the models were originally trained with perform the worse, while more explicit instructions work better.}
    \label{tab:big_models_prompt}
\end{table*}

\label{sec:prompts}
\subsection{Qwen2-VL-72B and InternVL2-76B Prompt Variants}
Our investigation reveals a surprising relationship between instruction format and bounding box prediction accuracy in large multimodal models (Qwen2-VL-72B and InternVL2-76B). Contrary to expectations, the default instructions employed during model training consistently yielded poorer performance. The model ignores the instruction and outputs a caption (see the attached examples in Tables~\ref{tab:results_qwen_org_prompt},~\ref{tab:results_qwen_prompt_1},~\ref{tab:results_qwen_prompt_2},~\ref{tab:results_qwen_prompt_3},~\ref{tab:results_internvl_org_prompt},~\ref{tab:results_internvl_prompt_1},~\ref{tab:results_internvl_prompt_2}, and~\ref{tab:results_internvl_prompt_3}). 
In contrast, more explicit and verbally detailed instructions produced significantly better results, and the outputs are indeed bounding boxes. 
In Table ~\ref{tab:big_models_prompt} we present a comprehensive analysis of various instruction formats and their corresponding performance outcomes across both models.
The best prompt over all datasets is reported in Table 2 of the main paper for each model.

\noindent We evaluated the models using four distinct prompting strategies, labeled Original prompt and prompt 1 through prompt 3: \\
\begin{figure}[htbp]
\fbox{%
\begin{minipage}{\linewidth}
Original prompt: \\
``\texttt{<ref>category</ref>}"
\end{minipage}%
}
\caption{Qwen2-VL Original Prompt}
\label{fig:org_prompt}
\end{figure}

\begin{figure}[htbp]
\fbox{%
\begin{minipage}{\linewidth}
Prompt 1: \\
``\texttt{Please provide the bounding box of the element \{element\}, return the bounding box in the following format: $[x_{0}, y_{0}, x_{1}, y_{1}]$}"
\end{minipage}%
}
\caption{Qwen2-VL / InternVL2 Prompt 1}
\label{fig:prompt1}
\end{figure}
\begin{figure}[htbp]
\fbox{%
\begin{minipage}{\linewidth}
Prompt 2: \\
``\texttt{Task: Locate the {element} in the image. Provide its bounding box coordinates in the format $[x_{min}, y_{min}, x_{max}, y_{max}]$}"
\end{minipage}%
}
\caption{Qwen2-VL / InternVL2 Prompt 2}
\label{fig:prompt2}
\end{figure}
\begin{figure}[htbp]
\fbox{%
\begin{minipage}{\linewidth}
Prompt 3: \\ 
``\texttt{Please analyze this image and locate the exact {element}. Return the precise bounding box coordinates using this format: $[x_{min}, y_{min}, x_{max}, y_{max}]$
The coordinates should tightly bound only the {element}, nothing more. Take your time to carefully examine the image and provide the most accurate bounding box possible.}"
\end{minipage}%
}
\caption{Qwen2-VL / InternVL2 Prompt 3}
\label{fig:prompt3}
\end{figure}

\subsubsection{Qwen2-VL-72B Examples:}
In Tables ~\ref{tab:results_qwen_org_prompt}, ~\ref{tab:results_qwen_prompt_1}, ~\ref{tab:results_qwen_prompt_2} and ~\ref{tab:results_qwen_prompt_3}, we provide additional and detailed examples of the variants of the prompts we used throughout our experiments.

\subsubsection{InternVL2-76B Examples:}
In Tables ~\ref{tab:results_internvl_org_prompt}, ~\ref{tab:results_internvl_prompt_1}, ~\ref{tab:results_internvl_prompt_2} and ~\ref{tab:results_internvl_prompt_3} we provide additional and detailed examples of the variants of the prompts we used throughout our experiments.

\subsection{GPT-4o Prompt Variants}
Following the successful experimental approaches implemented with Intern-VL2-76B and Qwen2-VL-72B, we employ the same instruction analysis for GPT-4o (gpt-4o-2024-08-06). 
We employ several prompt strategies to gain the best performance- Table \ref{tab:gpt_prompt} shows the full results over different prompts.
Concretely, we utilized the following prompt configurations ~\ref{fig:gpt_prompt1} and ~\ref{fig:gpt_prompt2}.

\begin{figure}[]
\fbox{%
\begin{minipage}{\linewidth}
``\texttt{Please provide the bounding box of the element {element}}"
\end{minipage}%
} 
\caption{GPT-4o Prompt 1}
\label{fig:gpt_prompt1}
\end{figure}
\begin{figure}    
\fbox{%
\begin{minipage}{\linewidth}
``\texttt{Please provide the bounding box of the element {element}, return the bounding box coordinates the following format: $[x_{min}, y_{min}, x_{max}, y_{max}]$. Do not output anything else besides the coordinate}"
\end{minipage}%
}
\caption{GPT-4o Prompt 2}
\label{fig:gpt_prompt2}
\end{figure}

\noindent In addition, we use the best resolution that the model can work with by setting the "detail: high" argument.
\noindent The best prompt over all datasets is reported in Table 2 of the main paper for each model.


\begin{table*}[hbtp]
    \centering
    \renewcommand{\arraystretch}{1.2}
    \setlength{\tabcolsep}{6pt}
    \resizebox{1.\textwidth}{!}{
\begin{tabular}{l l| c c| c c c c| c c c c| c}
    \toprule
    \multirow{2}{*}{\textbf{Model}} & \multirow{2}{*}{\textbf{Prompt}} & \multicolumn{2}{c|}{\textbf{PDM}} & \multicolumn{4}{c|}{\textbf{PerSeg}} & \multicolumn{4}{c|}{\textbf{ICL-LASOT}} & \multirow{2}{*}{\textbf{Average}} \\
    & & 1-shot & 2-shot & 1-shot & 2-shot & 3-shot & 4-shot & 1-shot & 2-shot & 4-shot & 8-shot & \\
    \midrule
    \midrule
    GPT-4o & Prompt 1 (~\ref{fig:gpt_prompt1}) & 4.83 &7.70 & 26.06 & 25.53 & 26.13 & 25.19 & 11.27 & 15.18 & 18.60 & 19.05 & 18.23 \\
    GPT-4o & Prompt 2 (~\ref{fig:gpt_prompt2}) & 11.27 & 13.29 & 33.43 & 41.41 & 44.77 & 45.72 & 18.13 & 20.61 & 23.08 & 24.61 & \textbf{27.63} \\
    \bottomrule
    \bottomrule
\end{tabular}}
    \caption{\textbf{Instruction format on GPT-4o.} We employ multiple prompt strategies to gain the best performance (\% IoU) out of the model.}
    \label{tab:gpt_prompt}    
\end{table*}
\subsubsection{GPT-4o Examples:}
In Tables ~\ref{tab:results_gpt_prompt_1} and ~\ref{tab:results_gpt_prompt_2} we provide additional and detailed examples of the variants of the prompts we used throughout our experiments.

\begin{table*}[htbp]
  \centering
  \begin{tabular}{p{8cm} | p{5cm} | c}
    \toprule
    \textbf{Prompt} & \textbf{Response} & \textbf{Visualization} \\
        \midrule
    \texttt{'<ref>balloon</ref>'}, '[456.25, 497.91, 517.18, 577.08]', \texttt{'<ref>balloon</ref>'} & 
    [444, 186, 506, 244] &
    \multirow{4}{*}{\includegraphics[width=3.5cm]{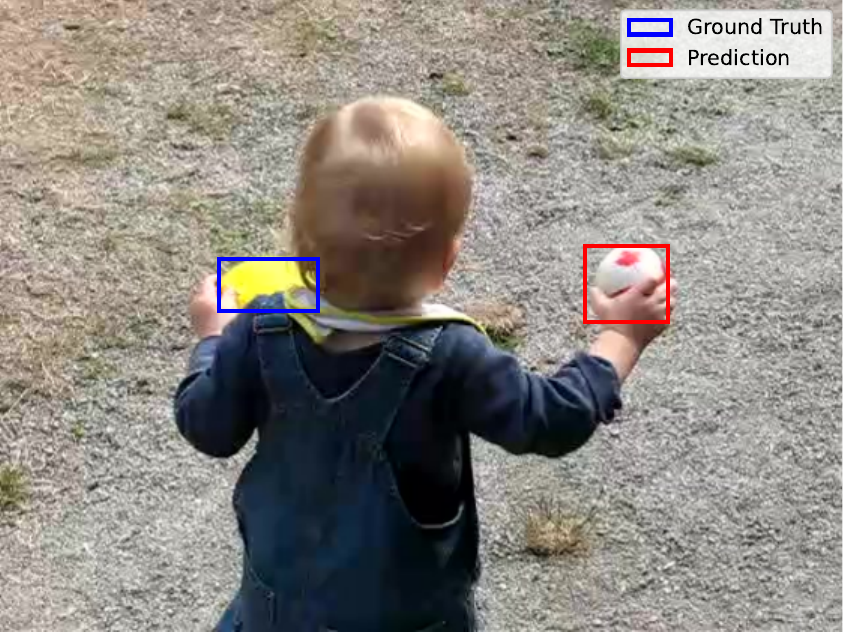}} \\
    \rule{0pt}{1.7cm} & & \\
    \midrule
     \texttt{'<ref>bear</ref>'}, '[519.53, 36.1, 606.25, 273.61]', \texttt{'<ref>bear</ref>'}, '[564.84, 320.83, 692.18, 577.77]', \texttt{'<ref>bear</ref>'} & 
     The image shows a group of bears walking along a dirt path in a forested area. The path is surrounded by trees and greenery. The bears appear to be moving in a line, with one bear following another. &
    \multirow{4}{*}{\includegraphics[width=3.5cm]{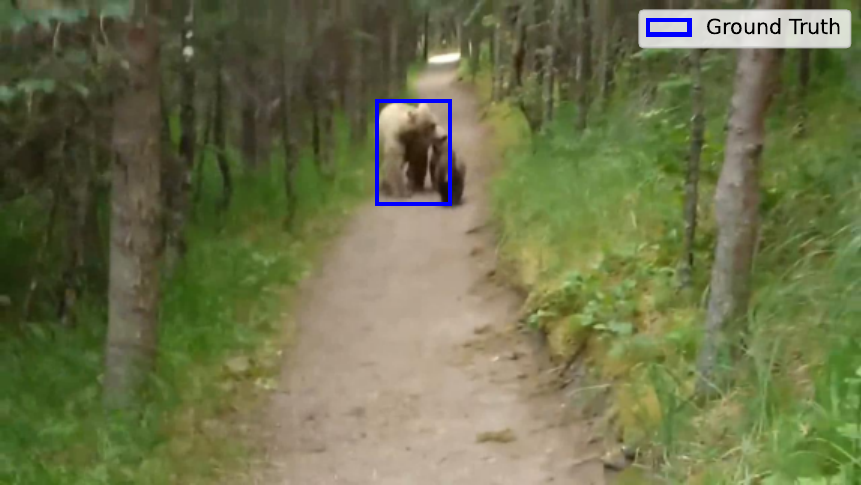}} \\
    \midrule
    \texttt{'<ref>dog</ref>'}, '[219.23, 56.41, 998.71, 926.92]', \texttt{'<ref>dog</ref>'}, '[120.35, 215.10, 998.71, 859.15]', \texttt{'<ref>dog</ref>'}, '[18.92, 257.64, 998.54, 951.96]', \texttt{'<ref>dog</ref>'} & 
    The image shows a fluffy dog with a white and orange coat, sitting and looking towards the camera. The dog has a happy expression with its tongue out. The background features a warm, orange hue with some blurred flowers, suggesting a pleasant, sunny environment. &
    \multirow{4}{*}{\includegraphics[width=3.5cm]{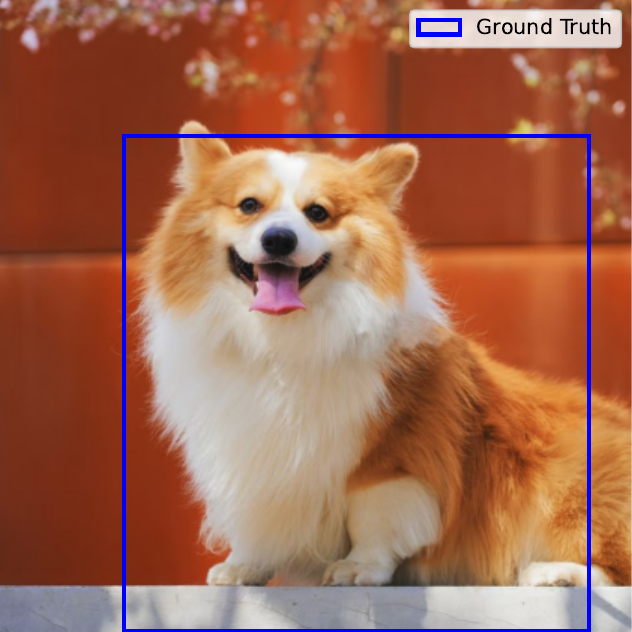}} \\ 
    \rule{0pt}{0.05cm} & & \\
    \bottomrule
  \end{tabular}
    \caption{\textbf{Qwen2-VL-72B with the original prompt examples.} We present examples from each dataset. Top row is 1-Shot from the PDM dataset, middle row is 2-shots from the ICL-LASOT test set and the bottom row is 3-shot from the PerSeg dataset}
  \label{tab:results_qwen_org_prompt}
\end{table*}

\begin{table*}[htbp]
  \centering
  \begin{tabular}{p{8cm} | p{5cm} | c}
    \toprule
    \textbf{Prompt} & \textbf{Response} & \textbf{Visualization} \\
    \midrule    
    'Please provide the bounding box of the element balloon, return the bounding box in the following format: [x0, y0, x1, y1]', '[456.25, 497.916, 517.1875, 577.083]', 'Please provide the bounding box of the element balloon, return the bounding box in the following format: [x0, y0, x1, y1]' & 
    [292, 239, 331, 276] &
    \multirow{4}{*}{\includegraphics[width=3.5cm]{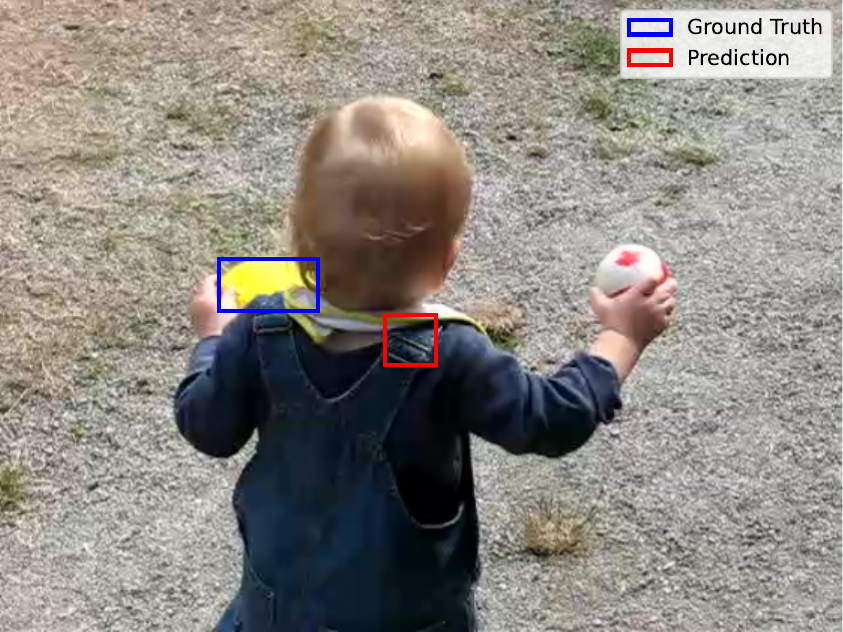}} \\
    \midrule
   'Please provide the bounding box of the element bear, return the bounding box in the following format: [x0, y0, x1, y1]', '[519.53, 36.11, 606.25, 273.61]', 'Please provide the bounding box of the element bear, return the bounding box in the following format: [x0, y0, x1, y1]', '[564.84, 320.833, 692.18, 577.77]', 'Please provide the bounding box of the element bear, return the bounding box in the following format: [x0, y0, x1, y1]' & 
    [641, 196, 694, 306] &
    \multirow{4}{*}{\includegraphics[width=3.5cm]{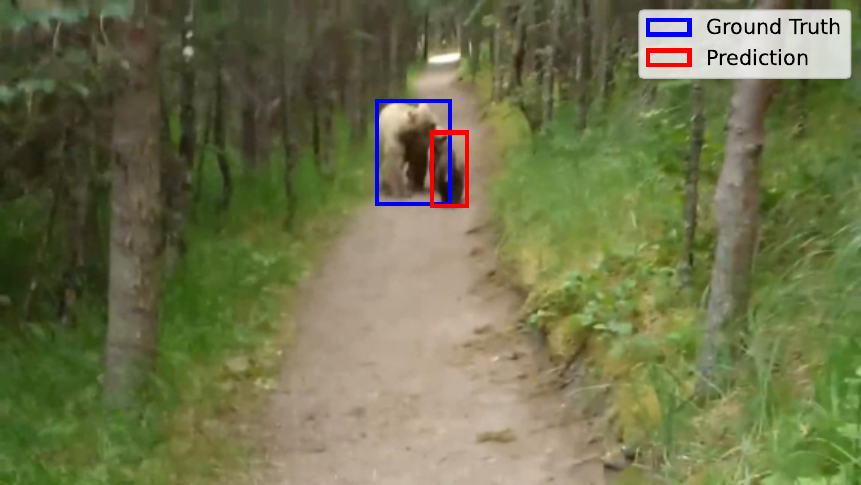}} \\
    \midrule
    'Please provide the bounding box of the element dog, return the bounding box in the following format: [x0, y0, x1, y1]', '[219.23, 56.41, 998.71, 926.92]', 'Please provide the bounding box of the element dog, return the bounding box in the following format: [x0, y0, x1, y1]', '[120.35, 215.10, 998.71, 859.15]', 'Please provide the bounding box of the element dog, return the bounding box in the following format: [x0, y0, x1, y1]', '[18.92, 257.64, 998.54, 951.96]', 'Please provide the bounding box of the element dog, return the bounding box in the following format: [x0, y0, x1, y1]' & 
    [171, 44, 779, 723] &
    \multirow{4}{*}{\includegraphics[width=3.5cm]{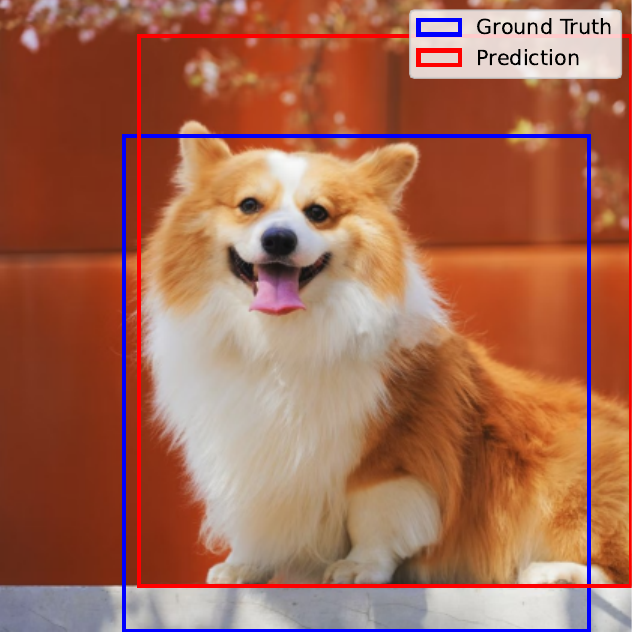}} \\
    \bottomrule
  \end{tabular}
    \caption{\textbf{Qwen2-VL-72B with prompt variant 1 examples.} We present examples from each dataset. Top row is 1-Shot from the PDM dataset, middle row is 2-shots from the ICL-LASOT test set and the bottom row is 3-shot from the PerSeg dataset}
  \label{tab:results_qwen_prompt_1}
\end{table*}

\begin{table*}[htbp]
  \centering
  \begin{tabular}{p{8cm} | p{5cm} | c}
    \toprule
    \textbf{Prompt} & \textbf{Response} & \textbf{Visualization} \\
    \midrule    
    'Task: Locate the balloon in the image. Provide its bounding box coordinates in the format [x\_min, y\_min, x\_max, y\_max]', '[456.25, 497.91, 517.18, 577.08]', 'Task: Locate the balloon in the image. Provide its bounding box coordinates in the format [x\_min, y\_min, x\_max, y\_max]' & 
    [199, 263, 222, 294] &
    \multirow{4}{*}{\includegraphics[width=3.5cm]{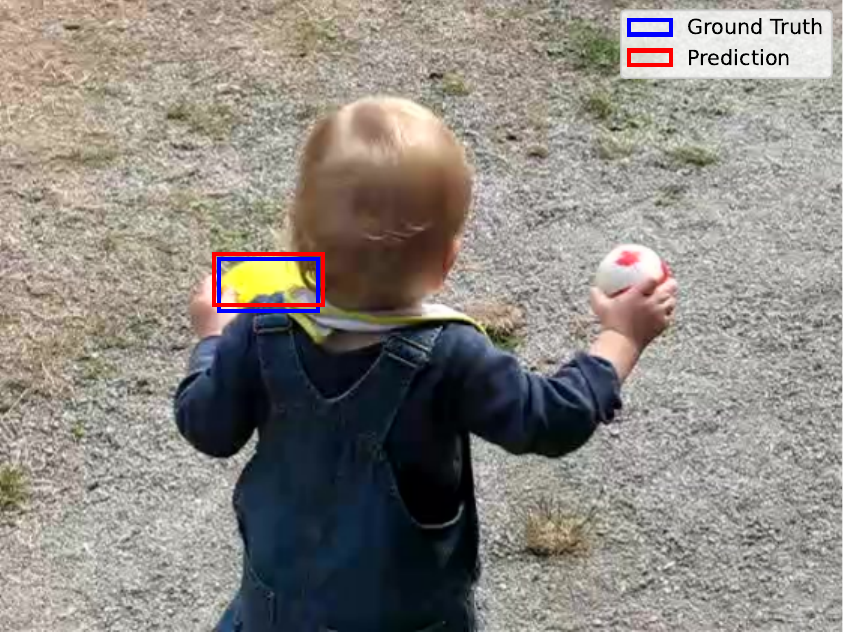}} \\
    \midrule
    'Task: Locate the bear in the image. Provide its bounding box coordinates in the format [x\_min, y\_min, x\_max, y\_max]', '[519.53, 36.11, 606.25, 273.61]', 'Task: Locate the bear in the image. Provide its bounding box coordinates in the format [x\_min, y\_min, x\_max, y\_max]', '[564.84, 320.8, 692.18, 577.77]', 'Task: Locate the bear in the image. Provide its bounding box coordinates in the format [x\_min, y\_min, x\_max, y\_max]' &
   [199, 263, 222, 294] &
    \multirow{4}{*}{\includegraphics[width=3.5cm]{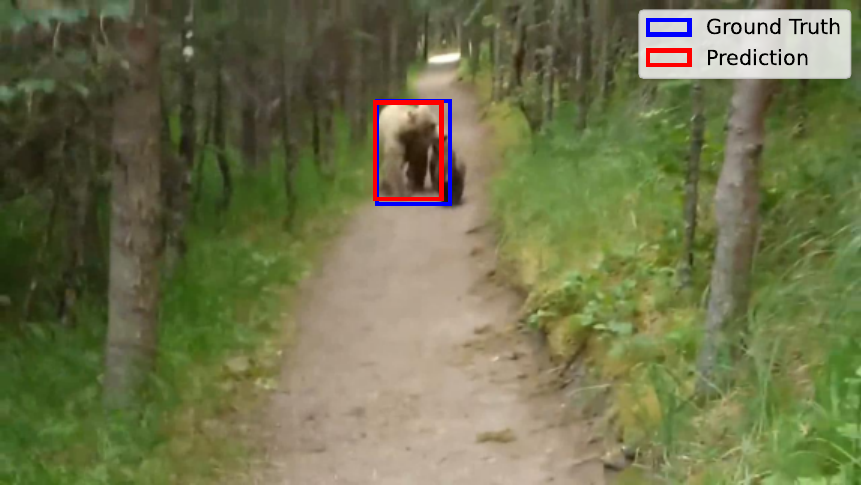}} \\
    \midrule
    'Task: Locate the dog in the image. Provide its bounding box coordinates in the format [x\_min, y\_min, x\_max, y\_max]', '[219.23, 56.4, 998.71, 926.92]', 'Task: Locate the dog in the image. Provide its bounding box coordinates in the format [x\_min, y\_min, x\_max, y\_max]', '[120.35, 215.10, 998.71, 859.15]', 'Task: Locate the dog in the image. Provide its bounding box coordinates in the format [x\_min, y\_min, x\_max, y\_max]', '[18.92, 257.64, 998.54, 951.96]', 'Task: Locate the dog in the image. Provide its bounding box coordinates in the format [x\_min, y\_min, x\_max, y\_max]' & 
    [151, 144, 779, 662] &
    \multirow{4}{*}{\includegraphics[width=3.5cm]{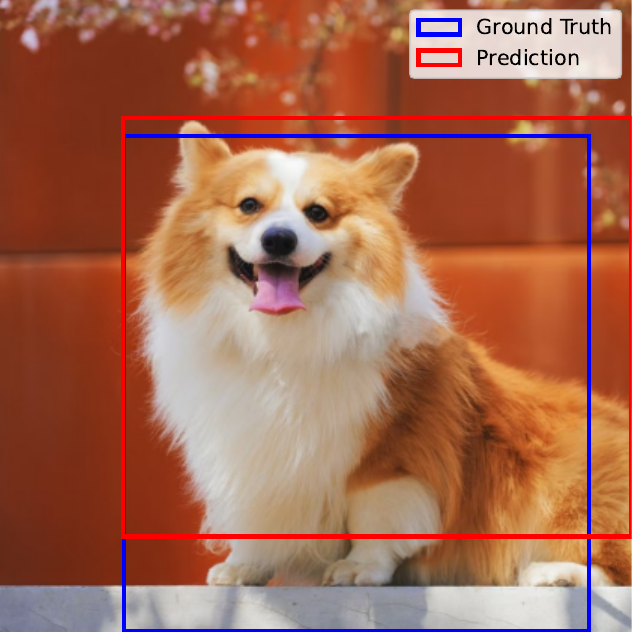}} \\
    \bottomrule
  \end{tabular}
    \caption{\textbf{Qwen2-VL-72B with prompt variant 2 examples.} We present examples from each dataset. Top row is 1-Shot from the PDM dataset, middle row is 2-shots from the ICL-LASOT test set and the bottom row is 3-shot from the PerSeg dataset.}
  \label{tab:results_qwen_prompt_2}
\end{table*}

\small
\begin{table*}[htbp]
  \centering
  \begin{tabular}{p{9cm} | p{3cm} | c}
    \toprule
    \textbf{Prompt} & \textbf{Response} & \textbf{Visualization} \\
    \midrule    
    'Please analyze this image and locate the exact balloon.  Return the precise bounding box coordinates using this format: [x\_min, y\_min, x\_max, y\_max] The coordinates should tightly bound only the balloon, nothing more Take your time to carefully examine the image and provide the most accurate bounding box possible.', '[456.25, 497.91, 517.18, 577.08]', 'Please analyze this image and locate the exact balloon.  Return the precise bounding box coordinates using this format: [x\_min, y\_min, x\_max, y\_max] The coordinates should tightly bound only the balloon, nothing more Take your time to carefully examine the image and provide the most accurate bounding box possible.' & 
    [0, 0, 0, 0] &
    \multirow{4}{*}{\includegraphics[width=3.5cm]{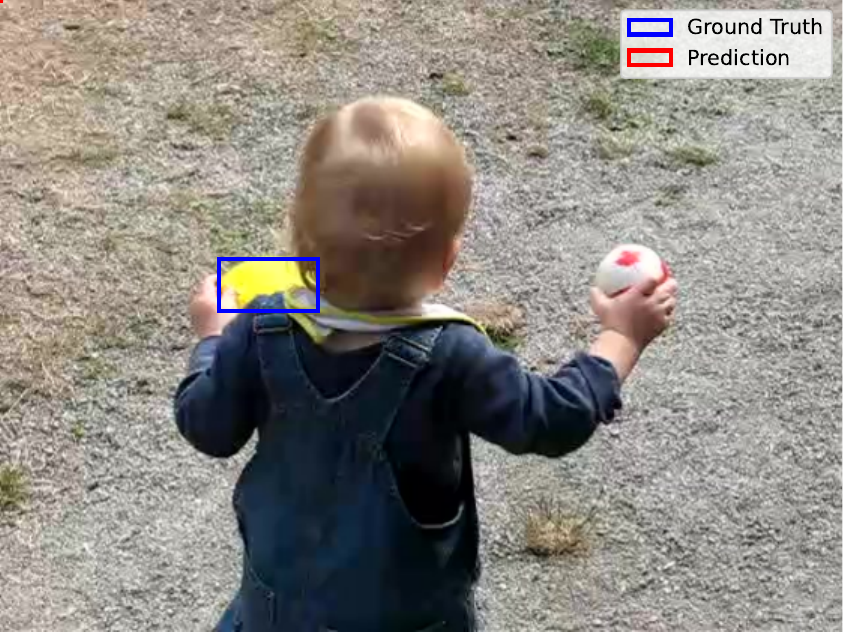}} \\
    \midrule
   'Please analyze this image and locate the exact bear.  Return the precise bounding box coordinates using this format:  [x\_min, y\_min, x\_max, y\_max] The coordinates should tightly bound only the bear, nothing more Take your time to carefully examine the image and provide the most accurate bounding box possible.', '[519.53, 36.11, 606.25, 273.61]', 'Please analyze this image and locate the exact bear.  Return the precise bounding box coordinates using this format:  [x\_min, y\_min, x\_max, y\_max] The coordinates should tightly bound only the bear, nothing more Take your time to carefully examine the image and provide the most accurate bounding box possible.', '[564.84, 320.83, 692.18, 577.77]', 'Please analyze this image and locate the exact bear.  Return the precise bounding box coordinates using this format:  [x\_min, y\_min, x\_max, y\_max] The coordinates should tightly bound only the bear, nothing more Take your time to carefully examine the image and provide the most accurate bounding box possible.' &
   [665, 25, 776, 197] &
    \multirow{4}{*}{\includegraphics[width=3.5cm]{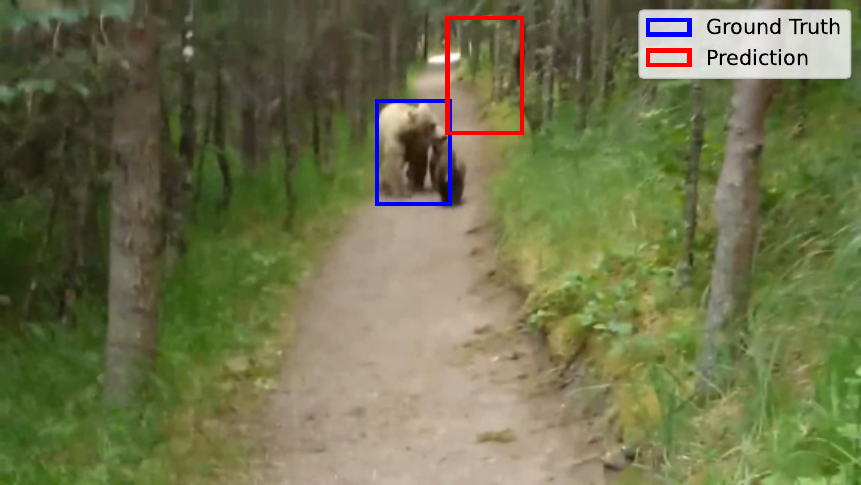}} \\
    \midrule
    'Please analyze this image and locate the exact dog.  Return the precise bounding box coordinates using this format: [x\_min, y\_min, x\_max, y\_max] The coordinates should tightly bound only the dog, nothing more Take your time to carefully examine the image and provide the most accurate bounding box possible.', '[219.230, 56.41, 998.71, 926.92]', 'Please analyze this image and locate the exact dog.  Return the precise bounding box coordinates using this format: [x\_min, y\_min, x\_max, y\_max] The coordinates should tightly bound only the dog, nothing more Take your time to carefully examine the image and provide the most accurate bounding box possible.', '[120.35, 215.10, 998.71, 859.15]', 'Please analyze this image and locate the exact dog.  Return the precise bounding box coordinates using this format: [x\_min, y\_min, x\_max, y\_max] The coordinates should tightly bound only the dog, nothing more Take your time to carefully examine the image and provide the most accurate bounding box possible.', '[18.92, 257.64, 998.54, 951.96]', 'Please analyze this image and locate the exact dog.  Return the precise bounding box coordinates using this format: [x\_min, y\_min, x\_max, y\_max] The coordinates should tightly bound only the dog, nothing more Take your time to carefully examine the image and provide the most accurate bounding box possible.' & 
    [138, 130, 780, 723] &
    \multirow{4}{*}{\includegraphics[width=3.5cm]{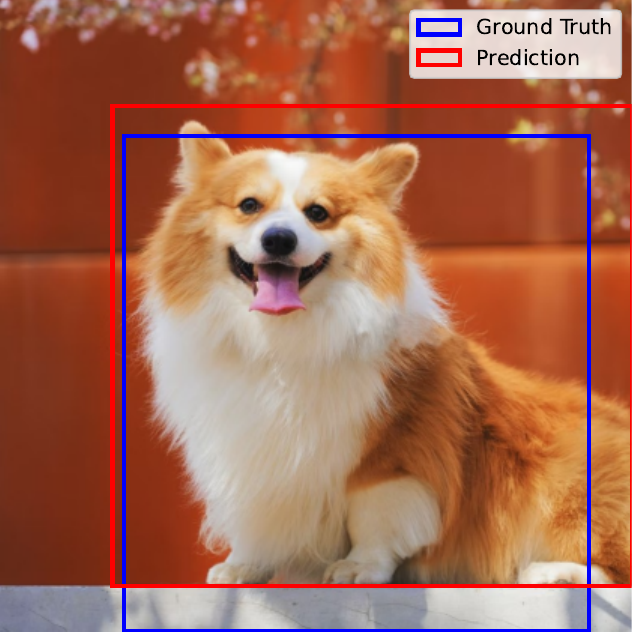}} \\
    \bottomrule
  \end{tabular}
    \caption{\textbf{Qwen2-VL-72B with prompt variant 3 examples.} We present examples from each dataset. Top row is 1-Shot from the PDM dataset, middle row is 2-shots from the ICL-LASOT test set and the bottom row is 3-shot from the PerSeg dataset.}
  \label{tab:results_qwen_prompt_3}
\end{table*}
\begin{table*}[htbp]
  \centering
  \begin{tabular}{p{8cm} | p{5cm} | c}
    \toprule
    \textbf{Prompt} & \textbf{Response} & \textbf{Visualization} \\
        \midrule
    ``\texttt{'<ref>balloon</ref>'}, '[456.25, 497.91, 517.18, 577.08]', \texttt{'<ref>balloon</ref>'}'' & 
    The second image shows a young child holding a yellow ball and a red and white ball, with one in each hand. The background is a gravel path with some green grass and a few scattered objects nearby. The child is wearing a blue shirt and blue overalls. &
    \multirow{4}{*}{\includegraphics[width=3.5cm]{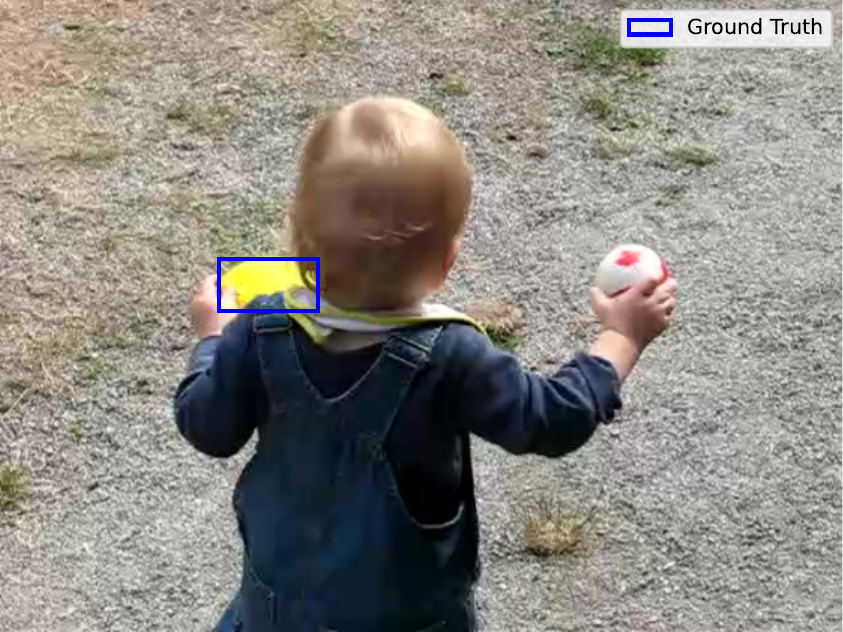}} \\    
    \midrule
     ``\texttt{'<ref>bear</ref>'}, '[519.53, 36.1, 606.25, 273.61]', \texttt{'<ref>bear</ref>'}, '[564.84, 320.83, 692.18, 577.77]', \texttt{'<ref>bear</ref>'}'' & 
     [558, 100, 679, 300] &
    \multirow{4}{*}{\includegraphics[width=3.5cm]{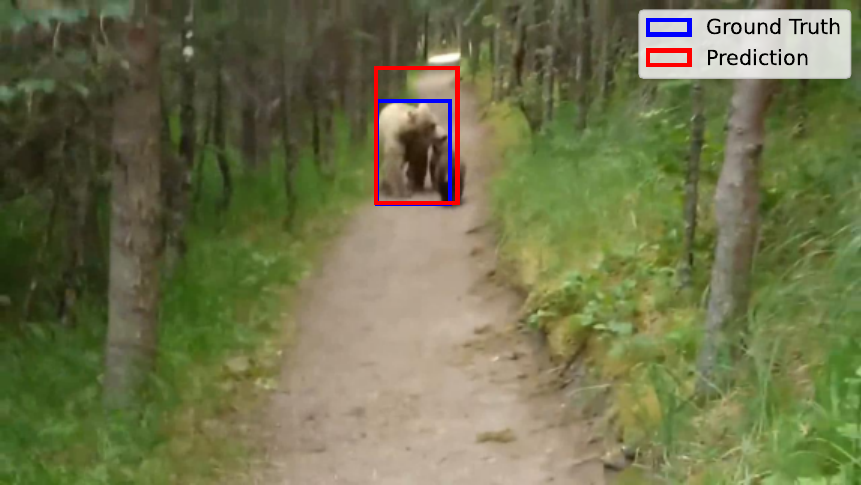}} \\
    \rule{0pt}{1.7cm} & & \\
    \midrule
    ``\texttt{'<ref>dog</ref>'}, '[219.23, 56.41, 998.71, 926.92]', \texttt{'<ref>dog</ref>'}, '[120.35, 215.10, 998.71, 859.15]', \texttt{'<ref>dog</ref>'}, '[18.92, 257.64, 998.54, 951.96]', \texttt{'<ref>dog</ref>'}'' & 
     [0, 149, 778, 728] &
    \multirow{4}{*}{\includegraphics[width=3.5cm]{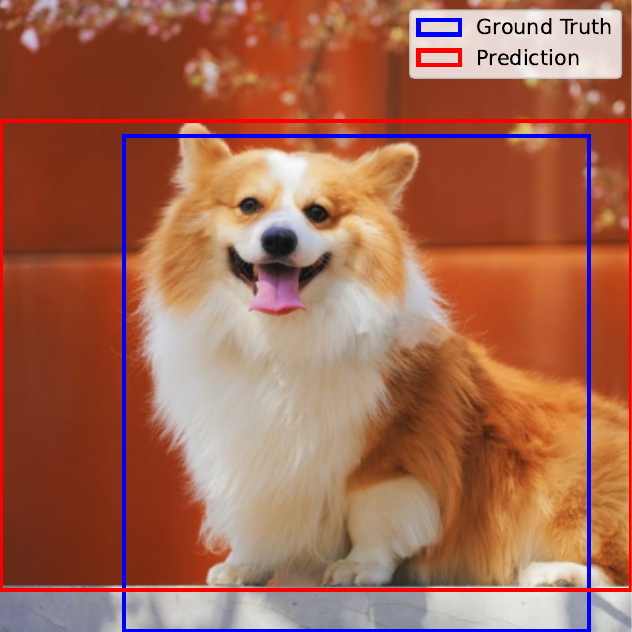}} \\ 
    \rule{0pt}{1.7cm} & & \\
    \bottomrule
  \end{tabular}
    \caption{\textbf{InternVL2-76B with the original prompt examples.} We present examples from each dataset. Top row is 1-Shot from the PDM dataset, middle row is 2-shots from the ICL-LASOT test set and the bottom row is 3-shot from the PerSeg dataset}
  \label{tab:results_internvl_org_prompt}
\end{table*}

\begin{table*}[htbp]
  \centering
  \begin{tabular}{p{8cm} | p{5cm} | c}
    \toprule
    \textbf{Prompt} & \textbf{Response} & \textbf{Visualization} \\
    \midrule    
    ``Please provide the bounding box of the element balloon, return the bounding box in the following format: [x0, y0, x1, y1]', '[456.25, 497.916, 517.18, 577.08]', 'Please provide the bounding box of the element balloon, return the bounding box in the following format: [x0, y0, x1, y1]'' & 
    [446, 187, 510, 247] &
    \multirow{4}{*}{\includegraphics[width=3.5cm]{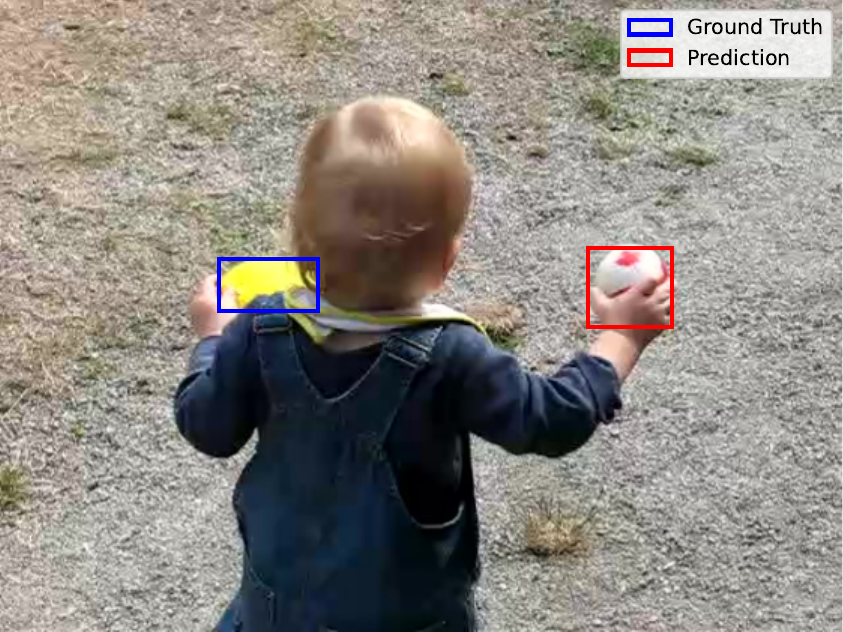}} \\
    \midrule
   ``Please provide the bounding box of the element bear, return the bounding box in the following format: [x0, y0, x1, y1]', '[519.53, 36.11, 606.25, 273.61]', 'Please provide the bounding box of the element bear, return the bounding box in the following format: [x0, y0, x1, y1]', '[564.84, 320.833, 692.18, 577.77]', 'Please provide the bounding box of the element bear, return the bounding box in the following format: [x0, y0, x1, y1]'' & 
    [595, 149, 689, 296] &
    \multirow{4}{*}{\includegraphics[width=3.5cm]{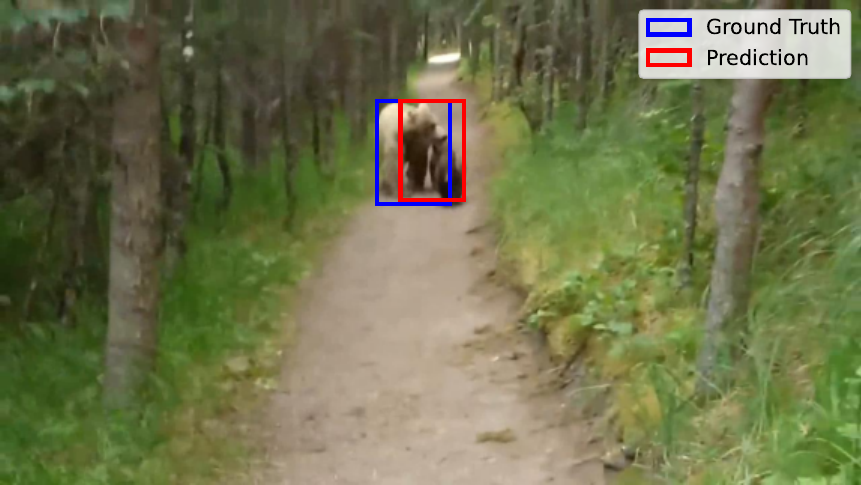}} \\
    \midrule
    `'Please provide the bounding box of the element dog, return the bounding box in the following format: [x0, y0, x1, y1]', '[219.23, 56.41, 998.71, 926.92]', 'Please provide the bounding box of the element dog, return the bounding box in the following format: [x0, y0, x1, y1]', '[120.35, 215.10, 998.71, 859.15]', 'Please provide the bounding box of the element dog, return the bounding box in the following format: [x0, y0, x1, y1]', '[18.92, 257.64, 998.54, 951.96]', 'Please provide the bounding box of the element dog, return the bounding box in the following format: [x0, y0, x1, y1]'' & 
    [103, 151, 780, 726]  &
    \multirow{4}{*}{\includegraphics[width=3.5cm]{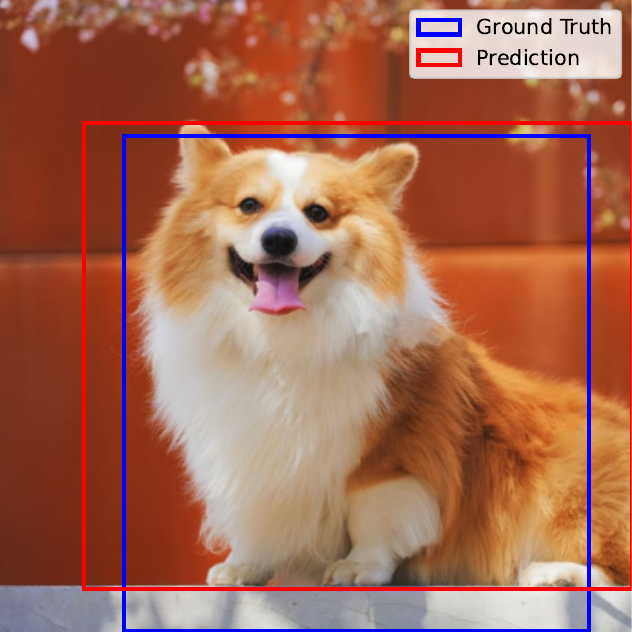}} \\
    \bottomrule
  \end{tabular}
    \caption{\textbf{InternVL2-76B with prompt variant 1 examples.} We present examples from each dataset. Top row is 1-Shot from the PDM dataset, middle row is 2-shots from the ICL-LASOT test set and the bottom row is 3-shot from the PerSeg dataset}
  \label{tab:results_internvl_prompt_1}
\end{table*}

\begin{table*}[htbp]
  \centering
  \begin{tabular}{p{8cm} | p{5cm} | c}
    \toprule
    \textbf{Prompt} & \textbf{Response} & \textbf{Visualization} \\
    \midrule    
    ``Task: Locate the balloon in the image. Provide its bounding box coordinates in the format [x\_min, y\_min, x\_max, y\_max]', '[456.25, 497.91, 517.18, 577.08]', 'Task: Locate the balloon in the image. Provide its bounding box coordinates in the format [x\_min, y\_min, x\_max, y\_max]'' & 
    [34, 159, 226, 218] &
    \multirow{4}{*}{\includegraphics[width=3.5cm]{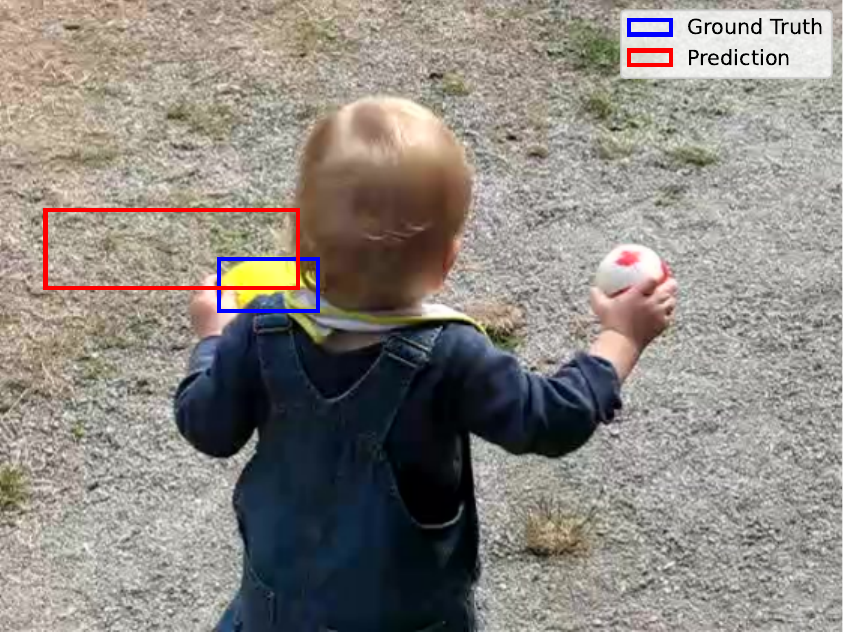}} \\
    \midrule
    ``Task: Locate the bear in the image. Provide its bounding box coordinates in the format [x\_min, y\_min, x\_max, y\_max]', '[519.53, 36.11, 606.25, 273.61]', 'Task: Locate the bear in the image. Provide its bounding box coordinates in the format [x\_min, y\_min, x\_max, y\_max]', '[564.84, 320.8, 692.18, 577.77]', 'Task: Locate the bear in the image. Provide its bounding box coordinates in the format [x\_min, y\_min, x\_max, y\_max]'' &
    560, 150, 688, 298] &
    \multirow{4}{*}{\includegraphics[width=3.5cm]{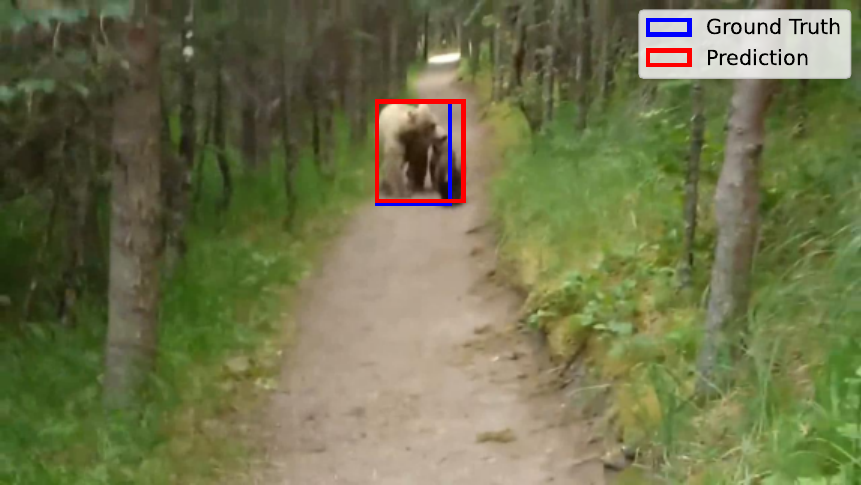}} \\
    \midrule
    ``Task: Locate the dog in the image. Provide its bounding box coordinates in the format [x\_min, y\_min, x\_max, y\_max]', '[219.23, 56.4, 998.71, 926.92]', 'Task: Locate the dog in the image. Provide its bounding box coordinates in the format [x\_min, y\_min, x\_max, y\_max]', '[120.35, 215.10, 998.71, 859.15]', 'Task: Locate the dog in the image. Provide its bounding box coordinates in the format [x\_min, y\_min, x\_max, y\_max]', '[18.92, 257.64, 998.54, 951.96]', 'Task: Locate the dog in the image. Provide its bounding box coordinates in the format [x\_min, y\_min, x\_max, y\_max]'' & 
   [162, 148, 778, 723] &
    \multirow{4}{*}{\includegraphics[width=3.5cm]{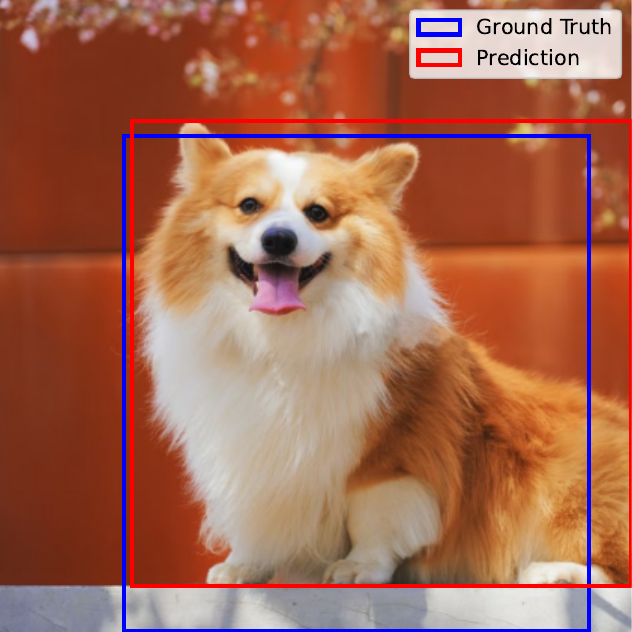}} \\
    \bottomrule
  \end{tabular}
    \caption{\textbf{InternVL2-76B with prompt variant 2 examples.} We present examples from each dataset. Top row is 1-Shot from the PDM dataset, middle row is 2-shots from the ICL-LASOT test set and the bottom row is 3-shot from the PerSeg dataset}
  \label{tab:results_internvl_prompt_2}
\end{table*}

\begin{table*}[htbp]
  \centering
  \begin{tabular}{p{9cm} | p{3cm} | c}
    \toprule
    \textbf{Prompt} & \textbf{Response} & \textbf{Visualization} \\
    \midrule    
    ``Please analyze this image and locate the exact balloon.  Return the precise bounding box coordinates using this format: [x\_min, y\_min, x\_max, y\_max] The coordinates should tightly bound only the balloon, nothing more Take your time to carefully examine the image and provide the most accurate bounding box possible.', '[456.25, 497.91, 517.18, 577.08]', 'Please analyze this image and locate the exact balloon.  Return the precise bounding box coordinates using this format: [x\_min, y\_min, x\_max, y\_max] The coordinates should tightly bound only the balloon, nothing more Take your time to carefully examine the image and provide the most accurate bounding box possible.'' & 
    446, 188, 509, 248] &
    \multirow{4}{*}{\includegraphics[width=3.5cm]{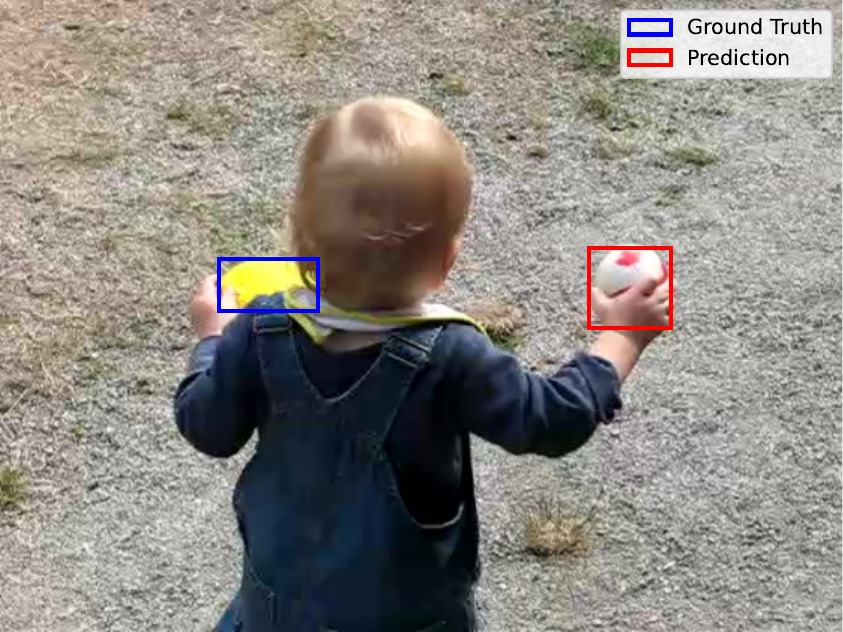}} \\
    \midrule
   ``Please analyze this image and locate the exact bear.  Return the precise bounding box coordinates using this format:  [x\_min, y\_min, x\_max, y\_max] The coordinates should tightly bound only the bear, nothing more Take your time to carefully examine the image and provide the most accurate bounding box possible.', '[519.53, 36.11, 606.25, 273.61]', 'Please analyze this image and locate the exact bear.  Return the precise bounding box coordinates using this format:  [x\_min, y\_min, x\_max, y\_max] The coordinates should tightly bound only the bear, nothing more Take your time to carefully examine the image and provide the most accurate bounding box possible.', '[564.84, 320.83, 692.18, 577.77]', 'Please analyze this image and locate the exact bear.  Return the precise bounding box coordinates using this format:  [x\_min, y\_min, x\_max, y\_max] The coordinates should tightly bound only the bear, nothing more Take your time to carefully examine the image and provide the most accurate bounding box possible.'' &
   [556, 149, 688, 295] &
    \multirow{4}{*}{\includegraphics[width=3.5cm]{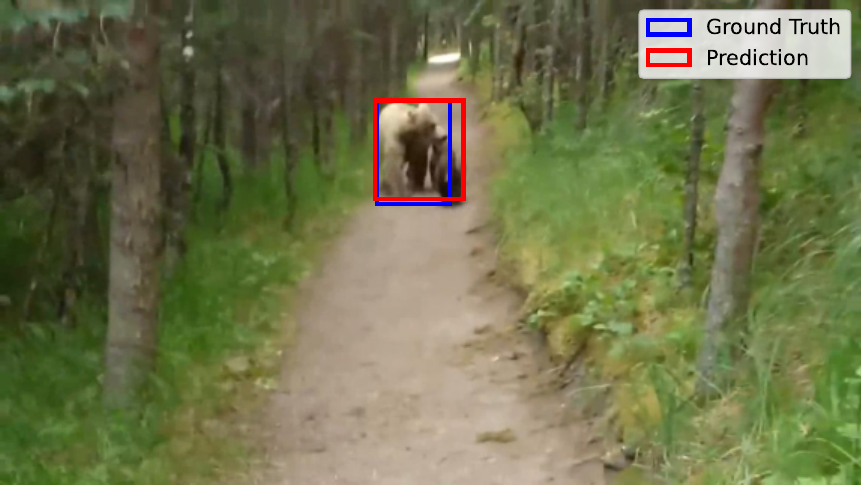}} \\
    \midrule
    ``Please analyze this image and locate the exact dog.  Return the precise bounding box coordinates using this format: [x\_min, y\_min, x\_max, y\_max] The coordinates should tightly bound only the dog, nothing more Take your time to carefully examine the image and provide the most accurate bounding box possible.', '[219.230, 56.41, 998.71, 926.92]', 'Please analyze this image and locate the exact dog.  Return the precise bounding box coordinates using this format: [x\_min, y\_min, x\_max, y\_max] The coordinates should tightly bound only the dog, nothing more Take your time to carefully examine the image and provide the most accurate bounding box possible.', '[120.35, 215.10, 998.71, 859.15]', 'Please analyze this image and locate the exact dog.  Return the precise bounding box coordinates using this format: [x\_min, y\_min, x\_max, y\_max] The coordinates should tightly bound only the dog, nothing more Take your time to carefully examine the image and provide the most accurate bounding box possible.', '[18.92, 257.64, 998.54, 951.96]', 'Please analyze this image and locate the exact dog.  Return the precise bounding box coordinates using this format: [x\_min, y\_min, x\_max, y\_max] The coordinates should tightly bound only the dog, nothing more Take your time to carefully examine the image and provide the most accurate bounding box possible.'' & 
    [150, 144, 778, 734] &
    \multirow{4}{*}{\includegraphics[width=4cm]{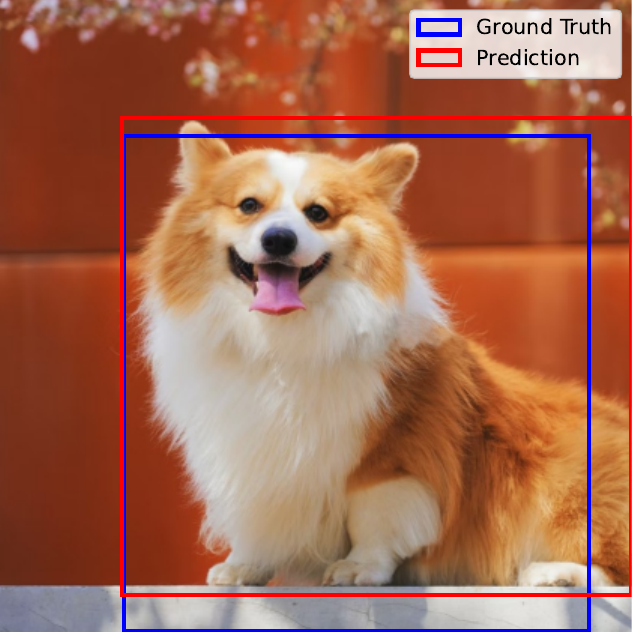}} \\
    \bottomrule
  \end{tabular}
    \caption{\textbf{InternVL2-76B with prompt variant 3 examples.} We present examples from each dataset. Top row is 1-Shot from the PDM dataset, middle row is 2-shots from the ICL-LASOT test set and the bottom row is 3-shot from the PerSeg dataset.}
  \label{tab:results_internvl_prompt_3}
\end{table*}
\begin{table*}[htbp]
  \centering
  \begin{tabular}{p{8cm} | p{5cm} | c}
    \toprule
    \textbf{Prompt} & \textbf{Response} & \textbf{Visualization} \\
    \midrule    
    ``Please provide the bounding box of the element balloon, '[456.25, 497.916, 517.18, 577.08]', 'Please provide the bounding box of the element balloon'' & 
The yellow balloon is not clearly visible. However, there is a yellow object in the child's left hand. Would you like details on that instead? &
    \multirow{4}{*}{\includegraphics[width=3.5cm]{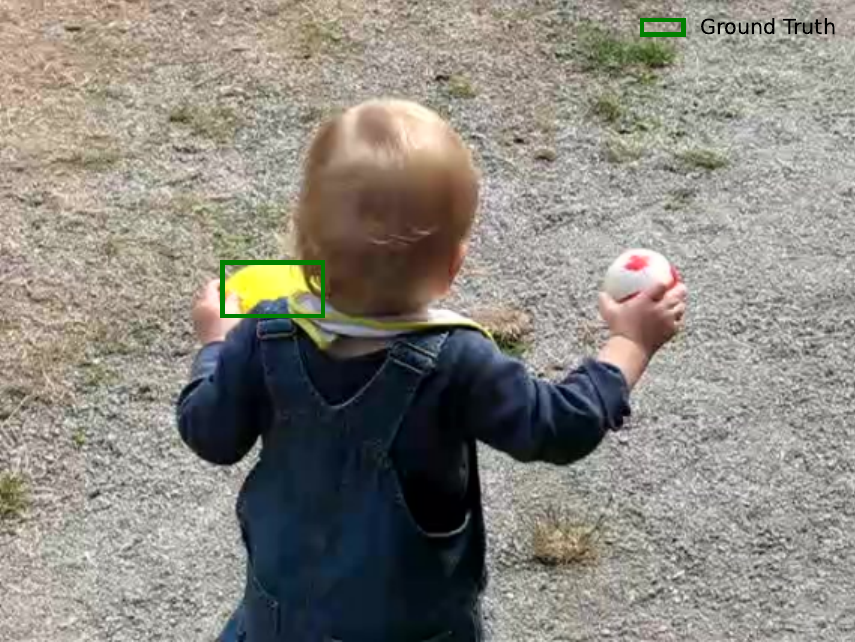}} \\
            \rule{0pt}{0.5cm} & & \\
    \midrule
   ``Please provide the bounding box of the element bear, '[519.53, 36.11, 606.25, 273.61]', 'Please provide the bounding box of the element bear, '[564.84, 320.833, 692.18, 577.77]', 'Please provide the bounding box of the element bear'' & 
    [531, 218, 700, 510] &
    \multirow{4}{*}{\includegraphics[width=3.5cm]{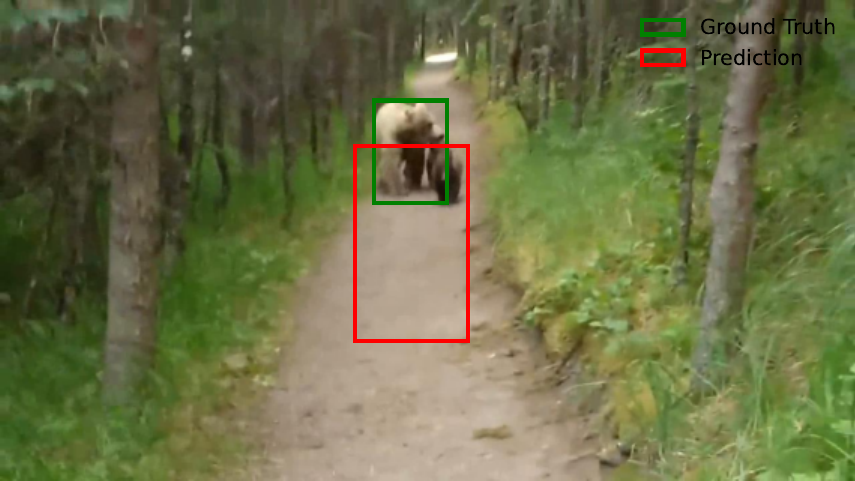}} \\
    \midrule
    ``Please provide the bounding box of the element dog, '[219.23, 56.41, 998.71, 926.92]', 'Please provide the bounding box of the element dog, '[120.35, 215.10, 998.71, 859.15]', 'Please provide the bounding box of the element dog, '[18.92, 257.64, 998.54, 951.96]', 'Please provide the bounding box of the element dog'' & 
     [167, 132, 616, 703] &
    \multirow{4}{*}{\includegraphics[width=3.5cm]{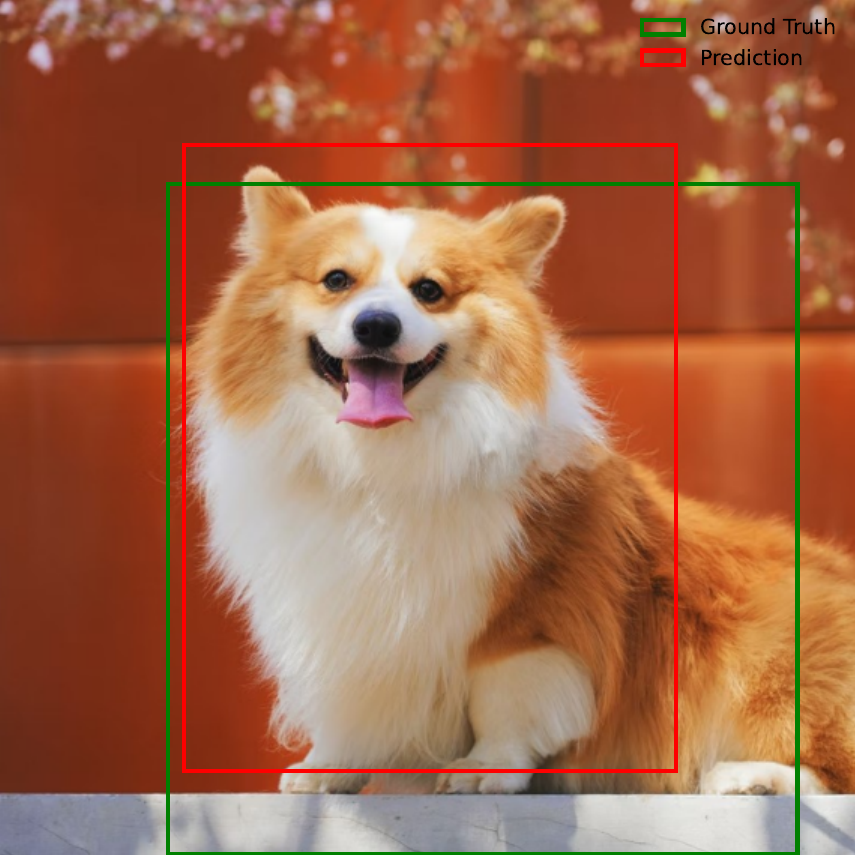}} \\
        \rule{0pt}{1.0cm} & & \\
    \bottomrule
  \end{tabular}
    \caption{\textbf{GPT-4o with prompt variant 1 examples.} We present examples from each dataset. Top row is 1-Shot from the PDM dataset, middle row is 2-shots from the ICL-LASOT test set and the bottom row is 3-shot from the PerSeg dataset}
  \label{tab:results_gpt_prompt_1}
\end{table*}

\begin{table*}[htbp]
  \centering
  \begin{tabular}{p{8cm} | p{5cm} | c}
    \toprule
    \textbf{Prompt} & \textbf{Response} & \textbf{Visualization} \\
    \midrule    
        ``Please provide the bounding box of the element balloon, return the bounding box coordinates the following format: $[x_{min}, y_{min}, x_{max}, y_{max}]$. Do not output anything else besides the coordinates", '[456.25, 497.916, 517.18, 577.08]', "Please provide the bounding box of the element balloon, return the bounding box coordinates the following format: $[x_{min}, y_{min}, x_{max}, y_{max}]$. Do not output anything else besides the coordinates'' & 
[190, 250, 250, 310] &
    \multirow{4}{*}{\includegraphics[width=3.5cm]{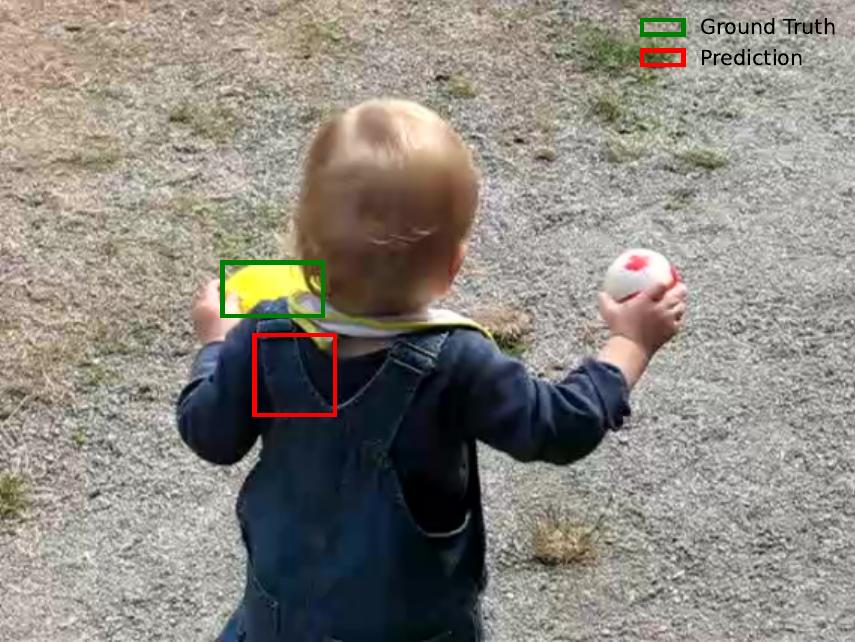}} \\
    \midrule
``Please provide the bounding box of the element bear, return the bounding box coordinates the following format: $[x_{min}, y_{min}, x_{max}, y_{max}]$. Do not output anything else besides the coordinates", '[519.53, 36.11, 606.25, 273.61]', "Please provide the bounding box of the element bear, return the bounding box coordinates the following format: $[x_{min}, y_{min}, x_{max}, y_{max}]$. Do not output anything else besides the coordinates", '[564.84, 320.833, 692.18, 577.77]', "Please provide the bounding box of the element bear, return the bounding box coordinates the following format: $[x_{min}, y_{min}, x_{max}, y_{max}]$. Do not output anything else besides the coordinates'' & 
    [576, 110, 694, 258] &
    \multirow{4}{*}{\includegraphics[width=3.5cm]{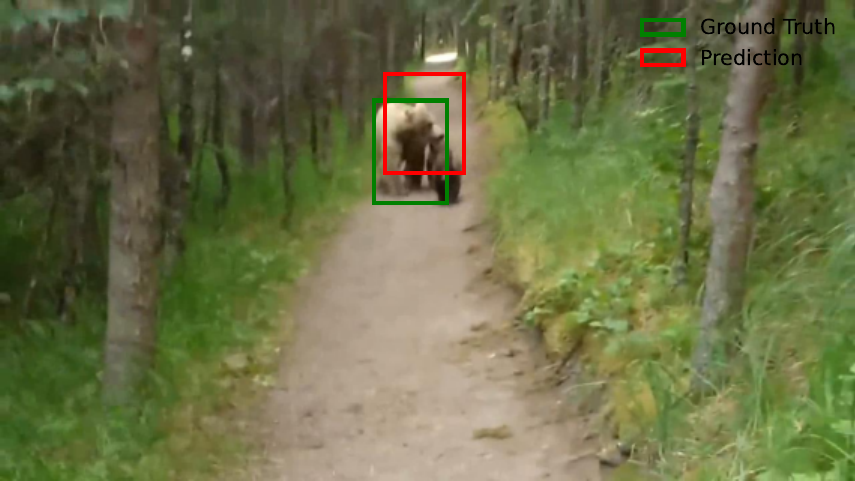}} \\
    \midrule
    ``Please provide the bounding box of the element dog, '[219.23, 56.41, 998.71, 926.92]', 'Please provide the bounding box of the element dog, '[120.35, 215.10, 998.71, 859.15]', 'Please provide the bounding box of the element dog, '[18.92, 257.64, 998.54, 951.96]', 'Please provide the bounding box of the element dog'' & 
     [167, 132, 616, 703] &
    \multirow{4}{*}{\includegraphics[width=3.5cm]{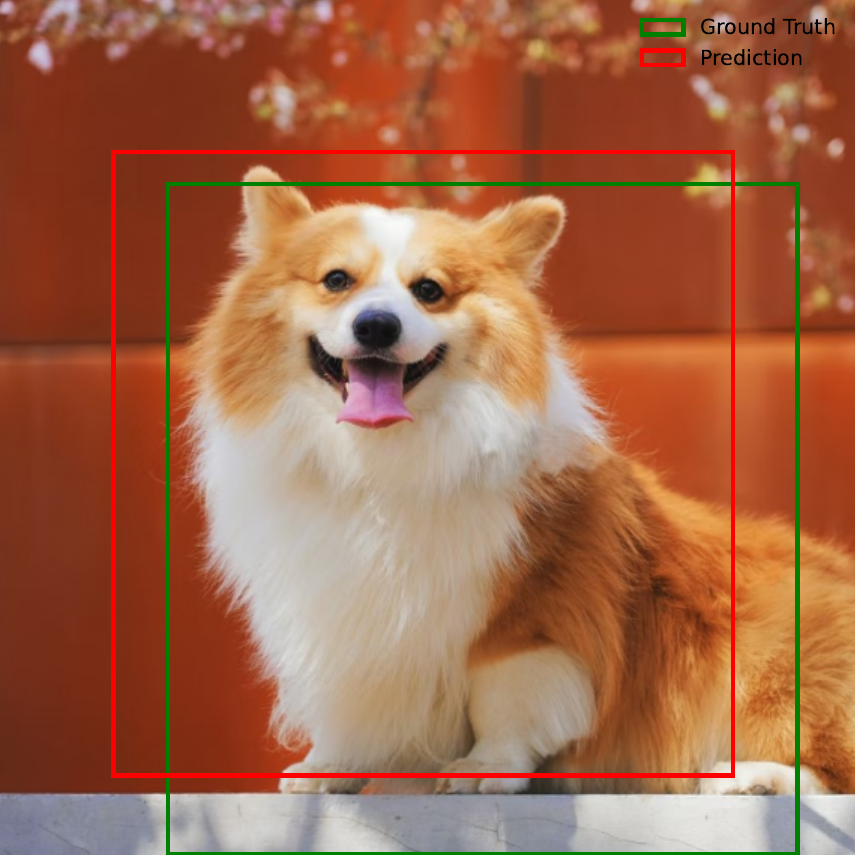}} \\
        \rule{0pt}{1.2cm} & & \\

    \bottomrule
  \end{tabular}
    \caption{\textbf{GPT-4o with prompt variant 2 examples.} We present examples from each dataset. Top row is 1-Shot from the PDM dataset, middle row is 2-shots from the ICL-LASOT test set and the bottom row is 3-shot from the PerSeg dataset}
  \label{tab:results_gpt_prompt_2}
\end{table*}

\end{document}


\begin{table}[h!]
\centering
\begin{tabular}{lcc}
\toprule
Metric        & Qwen2-VL-7B   & Our Model    \\
\midrule
seed\_1       & 0.7582 & 0.7504 \\
seed\_2       & 0.7979 & 0.7957 \\
seed\_3       & 0.7693 & 0.7606 \\
seed\_4       & 0.7022 & 0.7022 \\
seed\_5       & 0.7648 & 0.7554 \\
seed\_6       & 0.6720 & 0.6500 \\
seed\_7       & 0.7893 & 0.7906 \\
seed\_8       & 0.7484 & 0.7375 \\
seed\_9       & 0.7294 & 0.6588 \\
seed\_10      & 0.5991 & 0.5891 \\
seed\_image   & 0.7647 & 0.7574 \\
seed\_video   & 0.5515 & 0.5382 \\
seed\_all     & 0.7202 & 0.7116 \\

\bottomrule
\end{tabular}
\caption{Comparison of Qwen2-VL-7B and Our Model over SEED2 Benchmark splits}
\end{table}